\PassOptionsToPackage{unicode}{hyperref}
\PassOptionsToPackage{hyphens}{url}
\PassOptionsToPackage{dvipsnames,svgnames,x11names}{xcolor}
\documentclass[
  11pt,
]{article}
\usepackage{amsmath,amssymb}
\usepackage{iftex}
\ifPDFTeX
  \usepackage[T1]{fontenc}
  \usepackage[utf8]{inputenc}
  \usepackage{textcomp} 
\else 
  \usepackage{unicode-math} 
  \defaultfontfeatures{Scale=MatchLowercase}
  \defaultfontfeatures[\rmfamily]{Ligatures=TeX,Scale=1}
\fi
\usepackage{lmodern}
\ifPDFTeX\else
\fi
\IfFileExists{upquote.sty}{\usepackage{upquote}}{}
\IfFileExists{microtype.sty}{
  \usepackage[]{microtype}
  \UseMicrotypeSet[protrusion]{basicmath} 
}{}
\makeatletter
\@ifundefined{KOMAClassName}{
  \IfFileExists{parskip.sty}{%
    \usepackage{parskip}
  }{
    \setlength{\parindent}{0pt}
    \setlength{\parskip}{6pt plus 2pt minus 1pt}}
}{
  \KOMAoptions{parskip=half}}
\makeatother
\usepackage{xcolor}
\usepackage[margin=25mm]{geometry}
\usepackage{graphicx}
\makeatletter
\def\maxwidth{\ifdim\Gin@nat@width>\linewidth\linewidth\else\Gin@nat@width\fi}
\def\maxheight{\ifdim\Gin@nat@height>\textheight\textheight\else\Gin@nat@height\fi}
\makeatother
\setkeys{Gin}{width=\maxwidth,height=\maxheight,keepaspectratio}
\makeatletter
\def\fps@figure{htbp}
\makeatother
\ifLuaTeX
\usepackage[bidi=basic]{babel}
\else
\usepackage[bidi=default]{babel}
\fi
\babelprovide[main,import]{british}

\def\languageshorthands#1{}
\ifLuaTeX
  \usepackage{selnolig}  
\fi
\usepackage{bookmark}
\IfFileExists{xurl.sty}{\usepackage{xurl}}{} 
\hypersetup{
  pdftitle={Governing the Rule Space. Ruliology, Computational Irreducibility and Anticipatory Governance of Quantum Technologies},
  pdfauthor={Antonio Felice Uricchio; Francesco Scagliola},
  pdflang={en-GB},
  colorlinks=true,
  linkcolor={blue},
  filecolor={Maroon},
  citecolor={Blue},
  urlcolor={blue},
  pdfcreator={LaTeX via pandoc}}

\title{Governing the Rule Space. Ruliology, Computational Irreducibility
and Anticipatory Governance of Quantum Technologies}
\author{Antonio Felice Uricchio \and Francesco Scagliola}
\date{}

\begin{document}
\maketitle

A. F. Uricchio is Full Professor of Tax Law at the University of Bari
Aldo Moro, where he served as Rector (2013--2019). He is a former
President of ANVUR and currently a member of the Board of AVEPRO (Holy
See). F. Scagliola is Full Professor of Electroacoustic Music
Composition at the ``Niccolò Piccinni'' Conservatoire in Bari and
Coordinator of the PhD programme in Computational Electroacoustic Music
Composition, Generative Algorithmic Models and Computational Complexity.

\textbf{Original publication and permission.} This paper is an English
translation of the article originally published as: Antonio Felice
Uricchio and Francesco Scagliola, ``Governare lo spazio delle regole.
Ruliology, irriducibilità computazionale e governance anticipatoria
delle tecnologie quantistiche,'' Rivista della Corte dei conti, no. 4
(2026), pp.~27--41. The translation is distributed with the permission
of the journal.

\emph{Attribution of sections.} This article is the product of joint
reflection by the authors. While responsibility for its overall approach
and conclusions is shared, Sections 1, 2 and 8 are attributable to
Antonio Felice Uricchio; the remaining sections to Francesco Scagliola.

\textbf{CONTENTS:} 1. \emph{Future law}, political responsibility and
new technologies. -- 2. The limits of predictive rationality. -- 3. From
anticipation to exploration: a theoretical gap. -- 4. The Ruliological
Framework for Anticipatory Regulation. -- 5. Governing the European
quantum supply chain. -- 6. Three stress tests: openness and security,
standards, human capital. -- 7. Meta-rules, safeguards and limits. -- 8.
Conclusions.

\subsection{Abstract}\label{abstract}

The regulation of emerging technologies is often conceived as a strategy
of catching up, because legislators arrive too late. Yet gathering
signals, constructing scenarios and adapting rules to an anticipated
future are fundamental tasks for those entrusted with governance,
including the governance of rules themselves. The literature on
so-called \emph{future law} has already shown that anticipation means
building institutional capabilities rather than predicting outcomes
precisely (\emph{anticipatory governance}). Taking this premise as its
starting point, this article proposes to understand regulation as an
exploration of the transformations that may occur. The
\emph{Ruliological Framework for Anticipatory Regulation} distinguishes
rules, transformations and trajectories, assessing alternatives and
possible developments. Within this approach, computational
irreducibility is adopted conditionally, not as a demonstrated property
of legal systems. The \emph{Quantum Europe Strategy} provides a case
study and further grounds for reflection.

\textbf{Keywords:} \emph{future law}; ruliology; anticipatory
governance; computational irreducibility; regulatory robustness; quantum
technologies.

\subsection{\texorpdfstring{1. \emph{Future Law}, Political
Responsibility and New
Technologies}{1. Future Law, Political Responsibility and New Technologies}}\label{future-law-political-responsibility-and-new-technologies}

The expression \emph{Future Law} has appeared only recently in legal
discourse, reflecting its traditional focus on positive law and legal
history. Yet, in the face of the profound and rapid changes brought
about by new technologies,\footnote{See A. Contaldo, F. Dainotti,
  \emph{Diritto e tecnologie delle reti di informazione}, Edizioni
  Giuridiche Simone, Naples, 2005, p.~303; U. Draetta, \emph{Internet e
  commercio elettronico nel diritto internazionale dei privati},
  Giuffrè, Milan, 2001, p.~101; V. Frosini, \emph{Cibernetica, diritto e
  società}, Edizioni di Comunità, Milan, 1968; idem, \emph{Telematica e
  informatica giuridica}, in \emph{Enciclopedia del diritto}, vol.~XLIV,
  Giuffrè, Milan, 1992, p.~60; G. Pascuzzi, \emph{Il diritto dell'era
  digitale. Tecnologie informatiche e regole privatistiche}, Il Mulino,
  Bologna, 2002; F. Boccia, R. Leonardi (eds.), \emph{The Challenge of
  the Digital Economy: Markets, Taxation and Appropriate Economic
  Models}, Palgrave Macmillan, Cham, 2016, pp.~1 ff., DOI:
  10.1007/978-3-319-43690-6.} legal systems and legal scholarship must
move beyond a `defensive' or reactive approach,\footnote{G. ALPA,
  \emph{Prefazione}, in U. RUFFOLO (ed.), \emph{Intelligenza
  artificiale. Il diritto, i diritti, l'etica}, Milan, 2020, p.~XVII,
  who observes: `Although by its nature law follows the development of
  technology, and legal scholars sometimes intervene late in defining,
  classifying, ordering and regulating the new phenomena presented to
  them, this has not been the case with the revolution brought about by
  the application of artificial intelligence concepts {[}\ldots{]}
  Indeed, one might say that law, understood broadly as the ensemble of
  legislative, judicial and administrative acts, private activities and
  professional practices, has accompanied the development of new
  technologies step by step, studying their effects and promoting their
  development {[}\ldots{]} The fundamental dilemma is whether
  legislative intervention should be welcomed when a new phenomenon
  produces significant effects'.} offering regulatory and interpretative
responses capable of understanding, exploring and shaping the law of the
future, making it living and relevant to the present.

For too long, the regulation of emerging technologies has appeared slow
and unfocused, marked by a form of genuine double vision. Today, more
than ever, law must open itself to worlds that are no longer distant but
are forcefully entering socioeconomic reality: information technology,
cybernetics, robotics, biotechnology, artificial intelligence and
quantum technologies are not merely new and emerging scientific fields,
but phenomena with legal significance that must be considered,
interpreted and regulated from an evolutionary and forward-looking
perspective. Rather than a discipline, \emph{future law}\footnote{For
  various contributions on future law and the right to the future, see
  R. Bifulco, A. D'Aloia (eds.), \emph{Un diritto per il futuro. Teorie
  e modelli dello sviluppo sostenibile e della responsabilità
  intergenerazionale}, Jovene, Naples, 2008; M. A. Sandulli, F. Aperio
  Bella, L. Lorenzoni, N. Posteraro, M. Sinisi, \emph{Shaping the Future
  of Health Law: Challenges for Public Law}, paper published by
  \emph{federalismi.it}, 17 November 2021; T. J. Cottle, S. L.
  Klineberg, \emph{The Present of Things Future: Explorations of Time in
  Human Experience}, The Free Press, New York, 1974. The need to rethink
  modern law with regard to the rights of future generations is
  forcefully expressed by L. Tafaro, \emph{Diritti umani oggi: sviluppo
  sostenibile e generazioni future}, in A. A. Cançado Trindade, C.
  Barros Leal (eds.), \emph{Diritti umani e ambiente. VI Corso
  brasiliano interdisciplinare in diritti umani}, Fortaleza, 2017,
  pp.~43--73, especially p.~43, according to whom law is called upon to
  `ensure development centred on the human person, protect the rights
  (to existence and development) of present and future generations and
  (the survival of) the planet, bearing in mind that we are threatened
  not only by forces destructive of individuals and society, but also by
  forces destructive of the environment. Indeed, modern society seems
  more threatened by unsustainable development than by direct attacks on
  individual freedom'.} appears to be a working methodology that offers
legislators and legal interpreters regulatory avenues, including
experimental ones. Drawing on scientific evidence from outside law, it
connects with the rights of future generations, recently recognised in
the revised Articles 9 and 41 of the Italian Constitution.\footnote{See
  P. Torretta, \emph{Responsabilità intergenerazionale e procedimento
  legislativo. Soggetti, strumenti e procedure di positivizzazione degli
  interessi delle generazioni future}, in R. Bifulco, A. D'Aloia (eds.),
  \emph{Un diritto per il futuro}, cited above, pp.~699 ff.; M.
  Abrescia, \emph{Un diritto al futuro: analisi economica del diritto,
  Costituzione e responsabilità tra generazioni}, ibid., pp.~161 ff.; V.
  Pepe, \emph{Le «nuove frontiere» del costituzionalismo ambientale:
  prospettive emergenti nel diritto comparato}, in \emph{DPCE Online},
  vol.~58, special issue 2/2023, pp.~963 ff., DOI:
  10.57660/dpceonline.2023.1929, who argues that constitutional reform,
  while retaining an anthropocentric approach centred on the person,
  also indicates a biocentric perspective, calling for an objective and
  comprehensive conception of the environment as a context in which all
  living beings, animals and plants, have adequate space for life and
  social relations.} The breadth and comprehensiveness of these new
constitutional provisions highlight not only the prescriptive force of
the constitutional guarantee but also the responsibility placed on
present generations---and above all on political decision-makers---for
the benefit of future ones. This entails an entirely new conception of
the right to the future, hitherto addressed in legislation only in the
limited sense of children's right to development.\footnote{See, in
  particular, the Convention on the Rights of the Child, adopted by the
  United Nations General Assembly on 20 November 1989 and ratified by
  Italy through Law No.~176 of 27 May 1991; Article 6 guarantees every
  child the right to life, survival and development.} The United Nations
has adopted a considerably broader view through the so-called Pact for
the Future (New York, 22 September 2024), which also includes the
Declaration on Future Generations and a Global Digital Compact, laying,
in the words of the President of the General Assembly, `the foundations
for a sustainable, just and peaceful global order, for all peoples and
nations'. Through this commitment, participating countries sought to
accelerate progress on the major challenges of digital transition and
sustainable development and, more generally, on the human rights of
present and future generations. The close connection between the right
to the future, social sustainability and technological revolution calls
for a reconsideration of the scope of intergenerational duties of
solidarity and, consequently, of individual and collective
responsibilities.\footnote{See V. Pepe, \emph{Fare ambiente. Teorie e
  modelli giuridici di sviluppo sostenibile}, FrancoAngeli, Milan, 2008,
  according to whom the problem of sustainable development, as it has
  emerged in recent history, rests on the consideration that the
  opportunities of today's adult generation and those immediately
  following it to influence future generations globally, for better or
  worse, even those who will live in the distant future, appear vastly
  greater than those available to any preceding generation.
  Consequently, our responsibility towards posterity assumes greater
  importance than it could reasonably have had for earlier generations.
  The fundamental issue around which the ethical debate on future
  generations revolves is whether, assuming that those generations have
  rights, such rights---although belonging to individuals or groups who
  do not yet exist---can be legally enforced now, or whether future
  generations will have rights only to the resources existing at that
  later time.} The new constitutional provisions, together with the case
law of the European Court of Human Rights,\footnote{See also the case
  law of the European Court of Human Rights: judgment of 8 January 2009,
  application no. 12050/04, \emph{Mangouras v. Spain}; judgment of 27
  January 2009, application no. 67021/01, \emph{Tătar v. Romania};
  judgment of 10 January 2012, application no. 30765/08, \emph{Di Sarno
  and Others v. Italy}; judgment of 20 March 2008, application no.
  15339/02, \emph{Budayeva and Others v. Russia}.} give substance and
practical effect to the rights of present and future generations and to
the non-derogable duty of responsibility over time,\footnote{For
  approaches emphasising the deontic dimension of the principle and
  intergenerational responsibility, see G. Rivosecchi, \emph{La garanzia
  costituzionale della copertura finanziaria come vincolo intertemporale
  alla spesa pubblica nella (limitata) prospettiva della tutela delle
  generazioni future}, in R. Bifulco, A. D'Aloia (eds.), \emph{Un
  diritto per il futuro}, cited above, pp.~475 ff.; G. Arconzo, \emph{La
  sostenibilità delle prestazioni previdenziali e la prospettiva della
  solidarietà intergenerazionale. Al crocevia tra gli artt. 38, 81 e 97
  Cost.}, in \emph{Osservatorio costituzionale AIC}, issue 3/2018,
  pp.~627 ff.} now a new archetype to which legislation, the conduct of
institutions and individuals, public policies, private lifestyles,
production systems, businesses and workers must conform. Article 41 of
the Constitution subjects economic initiative to respect for the
environment, also in the interests of future generations, extending the
social and environmental purposes towards which it must be
directed.\footnote{See M. Greco, \emph{Il diritto costituzionale
  dell'ambiente dopo la riforma: alcune conferme e qualche (inattesa)
  novità nella sentenza della Corte costituzionale n.~105/2024}, in
  \emph{Diritti comparati}, 25 July 2024; E. Caterini, \emph{La
  «sostenibilità» nell'esperienza giuridica contemporanea. In occasione
  della discussione svoltasi nell'Ateneo barese}, in D. A. Benítez, C.
  Fava (eds.), \emph{Sostenibilità: sfida o presupposto?}, Milan, 2019,
  pp.~77 ff., especially p.~84, note 14; M. Pennasilico, review of E.
  Caterini, \emph{Sostenibilità e ordinamento civile. Per una
  riproposizione della questione sociale}, ESI, Naples, 2018, in
  \emph{Rassegna di diritto civile}, 2018, pp.~1511 ff., especially
  p.~1512; idem, \emph{La «sostenibilità ambientale» nella dimensione
  civil-costituzionale: verso un diritto dello «sviluppo umano ed
  ecologico»}, in \emph{Rivista quadrimestrale di diritto
  dell'ambiente}, no. 3/2020, pp.~4 ff.; D. Porena, \emph{Il principio
  della sostenibilità. Contributo allo studio di un programma
  costituzionale di solidarietà intergenerazionale}, Giappichelli,
  Turin, 2017, pp.~155 ff.}

In this changed context, digital technologies, artificial intelligence
systems and \emph{quantum computing} appear to be fundamental
instruments for giving practical effect to that responsibility and
enabling a better future, particularly when inspired by value systems
attentive to the person, including those not yet born, and oriented
towards the common good.\footnote{In this connection, consider the
  illuminating words of Pope Leo XIV's encyclical \emph{Magnifica
  Humanitas} on safeguarding the human person in the age of artificial
  intelligence and, in particular, on technological progress: `4. When
  Leo XIII spoke of ``new things'' (\emph{rerum novarum}), we cannot
  today merely repeat his precious teachings; rather, we must ask God
  for the wisdom to interpret the great trends of our time, especially
  technological advances. In recent years it has become increasingly
  evident how rapidly and profoundly digitalisation, artificial
  intelligence (AI) and robotics are transforming our world. Technology
  must not be regarded as a force inherently opposed to the human
  person: on the contrary, it has been rooted in our history from the
  beginning as a ``profoundly human reality, linked to human autonomy
  and freedom'' {[}5{]}. Over the centuries, technological development
  has contributed significantly to improving humanity's living
  conditions; at the same time, every phase of progress has revealed the
  ambiguous nature of tools capable of causing harm when not directed
  towards the good. Today, however, we face a new situation in which the
  power and pervasiveness of emerging technologies are woven into
  everyday life, shaping decision-making processes and profoundly
  influencing the collective imagination: ``Never has humanity had such
  power over itself'' {[}6{]}. New technologies open up a broad horizon
  in directions which, though we can glimpse them, we cannot yet fully
  predict. This makes it more difficult to assess their impact and
  long-term effects on human dignity and the common good. 5. It is now
  our turn to face the challenges of our time with clarity and
  responsibility. Appropriate legal instruments must be adopted, capable
  of protecting justice and containing the distorting effects of
  technological power. Yet the issue does not end with regulation. As
  Pope Francis warned, we must ask realistically who holds this power
  today and to what ends they direct it: ``We cannot ignore the fact
  that nuclear energy, biotechnology, information technology, knowledge
  of our own DNA and other capabilities we have acquired {[}\ldots{]}
  give those who possess knowledge, and especially the economic power to
  exploit it, an immense dominion over the whole human race and the
  entire world'' {[}7{]}. In the past, states were the principal actors
  guiding and directing innovation. Today, by contrast, its main drivers
  are private actors, often transnational, with resources and capacities
  for intervention greater than those of many governments. Technological
  power thus takes on a new, predominantly ``private'' form, making it
  even harder to discern, govern and direct towards the common good'.
  The bracketed numbers reproduce references to notes in the original:
  note {[}5{]} refers to Benedict XVI, encyclical \emph{Caritas in
  veritate} (29 June 2009), no. 69; note {[}6{]} refers to Francis,
  encyclical \emph{Laudato si'} (24 May 2015), no. 104; note {[}7{]},
  through `Ibid.', refers to that same source. The sign {[}\ldots{]}
  indicates an omission already present in the quotation reported by the
  encyclical.} Political decision-makers and legislators themselves are
called upon to promote innovation while also regulating it, anticipating
its effects, identifying risks and selecting, from among possible
options, those considered most suitable for pursuing shared values and
directed towards the good.\footnote{\emph{Magnifica Humanitas} also
  states: `Building a world in which everyone can ``flourish'' requires,
  thirdly, courageous shared responsibility. No single hand is
  sufficient to bear the weight of the challenges facing the world; nor
  is any hand so weak that it cannot make its contribution: ``For power
  is made perfect in weakness'' (2 Cor 12:9). Each person has their own
  part of the wall to build: scientists and researchers, entrepreneurs
  and workers, educators and legislators, civil society, popular
  movements and communities of faith. This is the logic of subsidiarity,
  which values cooperation between generations, peoples, disciplines and
  cultures as the principal path towards greater stability, prosperity
  and peace. Tensions and differences need not be feared: they can
  become creative energies when directed by shared responsibility'.}
When outcomes diverge from expectations and perceptions of unfairness
increase, responsibility is often attributed to political
decision-makers; yet the choices adopted sometimes rest on insufficient
data, imperfect models or untimely action. This gives rise to the need
for complete, analysed data, more extensive and rigorous
\emph{foresight}, more sensitive indicators and a more informed
legislative process capable of looking further ahead. In this regard,
Luigi Einaudi,\footnote{L. Einaudi, \emph{Conoscere per deliberare},
  first instalment of \emph{Prediche inutili}, Einaudi, Turin, 1956,
  pp.~1--12, where he writes: `The solution drags on'; `once a problem
  has been raised, it must be solved'; `it is urgent; we can delay
  addressing the issue no longer'. `Anyone reading such pronouncements
  wonders why the government, why Parliament, why the responsible
  minister are taking so long. Those who demand haste do not reflect: is
  this really the problem rather than merely one among many? How is it
  that there are so many urgent problems, different each day, whose
  solution cannot wait without harm, indeed without serious harm? Why is
  the list of urgent problems so long and the list of writings in which
  their substance is clearly explained so short? How can one decide
  without knowing? Yet nothing is more repugnant than knowledge to many,
  perhaps too many, of those called upon to solve problems'. He
  continues: `Is there any point in deciding without knowing? Decision
  must, in truth, be followed by action. One decides when one knows that
  action can be taken; decisions are not made to display barren wishful
  thinking. Yet nothing follows from a premature decision. {[}\ldots{]}
  Action also fails because the knowledge gathered with zealous
  enthusiasm has not infrequently lacked a guiding thread'. For other
  works by L. Einaudi, see \emph{Corso di scienza della finanza tenuto
  dal Prof.~Luigi Einaudi nella R. Università di Torino e nella
  Università commerciale L. Bocconi di Milano}, La Riforma Sociale,
  Turin, 1916; \emph{Il sistema tributario italiano}, 2nd ed., La
  Riforma Sociale, Turin, 1933; \emph{Miti e paradossi della giustizia
  tributaria}, Einaudi, Turin, 1940; \emph{Saggi sul risparmio e
  l'imposta}, Einaudi, Turin, 1941; \emph{La terra e l'imposta},
  Einaudi, Turin, 1942.} in his \emph{Prediche inutili}, formulated the
celebrated maxim `first acquire knowledge, then discuss, then decide',
emphasising that any financial, economic or political decision should be
preceded by the acquisition of a rich, comprehensive and appropriate
body of information, avoiding hasty and insufficiently considered
measures. The Turin economist, later the second President of the Italian
Republic, also observed that `it is not the one who searches who knows,
but the one who knows how to search', highlighting the value of
knowledge and research within the necessary relationship between
institutions, businesses and civil society, also by drawing on the
extraordinary potential of new technologies such as \emph{big data},
artificial intelligence and quantum technologies.

The OECD report \emph{Governing with Artificial Intelligence}, approved
by the Public Governance Committee on 5 September 2025,\footnote{OECD,
  \emph{Governing with Artificial Intelligence: The State of Play and
  Way Forward in Core Government Functions}, OECD Publishing, Paris,
  2025, DOI: 10.1787/795de142-en.} also points in this direction. By
examining two hundred uses of artificial intelligence across eleven core
government functions, it shows that enhancing public decision-makers'
knowledge capabilities---from forecasting fiscal risks to combating
fraud---is now recognised as a decisive factor in the effectiveness and
legitimacy of administrative action. No less significant is the finding
in the OECD's \emph{Tax Administration 2025}\footnote{OECD, \emph{Tax
  Administration 2025: Comparative Information on OECD and Other
  Advanced and Emerging Economies}, OECD Publishing, Paris, 2025, DOI:
  10.1787/cc015ce8-en.} that the proportion of tax administrations using
artificial intelligence rose from 9\% in 2016 to 69\% in 2023, further
indicating that the traditional Italian intuition---knowing in order to
tax, taxing in order to govern---is finding practical expression in
contemporary technology, as the founding figures of the discipline had
perceptively anticipated.

It follows that new technologies and complete, appropriate information
resources can improve the quality of public action, foster better
relationships between citizens and institutions through greater mutual
trust, and pursue the common good and the public interest more
effectively.

\subsection{2. The Limits of Predictive
Rationality}\label{the-limits-of-predictive-rationality}

As the \emph{Report on Intelligence Policy for Security}, submitted
annually to Parliament by the Italian Government, points out, the
capacity for `strategic anticipation' depends on the quality of the
State's knowledge infrastructure,\footnote{Presidency of the Council of
  Ministers, Intelligence System for the Security of the Republic,
  \emph{Relazione sulla politica dell'informazione per la sicurezza
  relativa all'anno 2024}, Rome, 2025.} which is itself sustained by the
capital of the new millennium: data made available, collected and
processed.

The OECD, in \emph{The Path to Becoming a Data-Driven Public Sector}
(2019), likewise emphasised that governments must place the role and
value of data at the centre of social and political transformation,
recognising data as an integral strategic asset for policymaking,
service delivery and institutional innovation.\footnote{OECD, \emph{The
  Path to Becoming a Data-Driven Public Sector}, OECD Digital Government
  Studies, OECD Publishing, Paris, 2019, DOI: 10.1787/059814a7-en. For
  the wording quoted subsequently, see C. van Ooijen, B. Ubaldi, B.
  Welby, \emph{A data-driven public sector: Enabling the strategic use
  of data for productive, inclusive and trustworthy governance}, OECD
  Working Papers on Public Governance, no. 33, OECD Publishing, Paris,
  2019, DOI: 10.1787/09ab162c-en, abstract on the OECD publisher's page:
  \emph{governments now need to go further and put the role and value of
  data at the core of thinking about the digital transformation of
  government}.} It should not be overlooked, however, that---as Michel
Foucault cautioned---the relationship between knowledge and power, the
celebrated \emph{pouvoir/savoir} nexus, is never neutral: whoever
controls flows of knowledge controls structures of power. It follows
that decision-makers must have access to, and control over, data with
full transparency, as a condition of democratic governance.

More generally, public decision-making models have historically
developed around three fundamental poles: will, power and
knowledge.\footnote{See M. Palumbo, \emph{Il processo di valutazione.
  Decidere, programmare, valutare}, FrancoAngeli, Milan, 2001. On this
  subject, see A. F. De Toni, \emph{Prefazione}, in P. Adinolfi, A. F.
  Uricchio (eds.), \emph{Valutare la complessità. La performance nelle
  organizzazioni pubbliche}, Carocci, Rome, 2025, p.~15; in the same
  volume, G. Fattore, \emph{La valutazione delle politiche: una breve
  guida metodologica}, pp.~179 ff., according to whom `ongoing and
  ex-post evaluations require field data on changes brought about by
  policies. This calls for both the willingness and capacity to collect
  data, either \emph{ad hoc} (designed specifically to measure policy
  effects) or drawn from existing information flows'. The author further
  observes that `every policy should be prepared on the basis of a
  systematic ex-ante assessment and should include plans for designing
  ongoing and ex-post evaluations'.} Experience shows that these
criteria alone are insufficient: public choices must also be capable of
generating collective well-being, equity and inclusion.

From this perspective, law is not the crystallisation of a causal order
but the normative expression of a purposive order inspired by the public
interest and the common good. The capacity to act within the field of
available choices depends on knowledge, understood not as contemplative
learning but as an active function mediating between the complexity of
reality and the urgency of choice. Einaudi's injunction to `know' is not
a preparatory moment that ends before a decision; it is a permanent
disposition, an institutional \emph{habitus} that must permeate the
entire deliberative cycle, from framing the problem to evaluating data
and outcomes. Without this discipline of knowledge, public deliberation
remains exposed to the dual risks of ideologism---deciding according to
what one believes---and decisionism---deciding simply because one has
the power to do so. The strength of an effective public decision lies
instead in the network of values that inspires it and, consequently, in
the interest that guides it. That interest cannot be individual: it must
necessarily transcend the individual and thus be public, concerning both
the present and the future. In this perspective, knowledge and data
analysis are central, also through new technologies. In 1597, Francis
Bacon coined the expression \emph{ipsa scientia potestas est},
`knowledge itself is power', in his \emph{Meditationes
Sacrae},\footnote{F. Bacon, \emph{Meditationes Sacrae} (1597), in J.
  Spedding, R. L. Ellis, D. D. Heath (eds.), \emph{The Works of Francis
  Bacon}, London, 1857--1874. The formulation \emph{scientia potentia
  est} appears in the Latin version of T. Hobbes's \emph{Leviathan},
  published in 1668.} anticipating the centrality of knowledge as the
foundation of public action. This insight must not be reduced to a logic
of domination: the knowledge in question is not power for its own sake,
nor power founded on force or will, but power deriving from data,
analysis and reflection. Applied to the present, knowing in order to
decide thus means acquiring data, but above all verifying them, because
decision-making must be informed and grounded in
self-determination---something now placed at risk when data are
contaminated (\emph{data pollution}).

The OECD's \emph{Digital Government Index} (2023) found that, although
member countries had made significant progress towards data-driven
public sectors, with South Korea, Estonia, Ireland, Norway and Denmark
among the most advanced, maturity in \emph{data governance} remained
uneven and only 15.8\% of countries had policy instruments (such as
laws, standards and guidelines) to help public institutions explain how
and why they use algorithmic tools.\footnote{OECD, \emph{2023 OECD
  Digital Government Index: Results and Key Findings}, OECD Public
  Governance Policy Papers, no. 44, OECD Publishing, Paris, 2024, DOI:
  10.1787/1a89ed5e-en. The countries most advanced in the `Data-driven
  public sector' dimension were South Korea, Estonia, Ireland, Norway,
  Denmark, France, Colombia, Australia, Lithuania and Sweden.} This
finding shows how much remains to be done to achieve full algorithmic
transparency in the public sector.

Against this background, several systemic challenges must be identified
and addressed in decision-making processes. The first is
\textbf{uncertainty}, concerning what is unknown or cannot be reliably
estimated, including everything that has not been verified. The second
is \textbf{complexity}, concerning not only the set of data but also the
relationships among elements and the interactions among social groups,
organisations and businesses, which are far from linear and often depend
on the context in which they operate. The third is \textbf{computational
irreducibility}, understood as a property of certain processes for which
no general procedure is available to determine the outcome without
following, at least in its essential steps, the process's
evolution.\footnote{S. WOLFRAM, \emph{A New Kind of Science}, Champaign,
  2002, pp.~737--742.} Determining the state reached by a system may
require executing or simulating intermediate states. Computational
irreducibility is equivalent neither to randomness nor to absolute
unpredictability. Individual outcomes may be anticipated, some
properties may be calculated, and particular classes of processes may
admit simplifications. The claim concerns the absence of a general
predictive shortcut applicable to the process's entire evolution.

\subsection{3. From Anticipation to Exploration: A Theoretical
Gap}\label{from-anticipation-to-exploration-a-theoretical-gap}

In light of the foregoing, more information may reduce uncertainty and
provide valuable support for decision-making, but it does not
necessarily eliminate complexity or guarantee that every dynamic can be
compressed into a reliable prediction. As Israeli and Goldenfeld have
shown with reference to cellular automata, processes that are
computationally irreducible at a fine scale may admit predictive or
computationally reducible descriptions through \emph{coarse-graining},
that is, a representation that aggregates states and disregards
microscopic details irrelevant to the property under
observation.\footnote{N. ISRAELI, N. GOLDENFELD, \emph{Computational
  Irreducibility and the Predictability of Complex Physical Systems}, in
  \emph{Physical Review Letters}, 2004, vol.~92, art. 074105, DOI:
  10.1103/PhysRevLett.92.074105.} The irreducibility of an entire
evolution does not therefore exclude the predictability of certain
aggregate quantities. Within the RFAR, this caution is expressed by the
function \(\pi_{P,S,c,h}\), introduced below, which selects a property
\(P\), a state, a context and a horizon. The framework does not call for
abandoning prediction where an appropriate reduction is available; it
requires the level of description to be specified, the properties that
remain predictable to be identified, and regulatory exploration to be
used when reduction loses information crucial to public choice.

Transferring this notion to law requires caution. One cannot maintain
that a legal system is an automaton, that social behaviour is fully
computable, or that the irreducibility of regulatory ecosystems has been
demonstrated. The notion is adopted as a methodological hypothesis: if,
for some sociotechnical transformations, there is no sufficiently
reliable predictive shortcut, a regulatory rationality based exclusively
on prediction is fragile. Regulators must then be able to learn also
from the controlled application of alternatives, the observation of
transformations and the revision of rules.

This conclusion does not entail abandoning rationality. It requires
changing its object. The question is no longer only `what future will
materialise?', but `which rules can we adopt, which transformations
might they produce, which trajectories might they open or close, and
which conditions must remain protected?'. The first question seeks a
privileged description of the future. The second establishes a
discipline of exploration. The distinction is particularly important for
quantum technologies, whose components mature at different rates.
Subsidies, procurement and standards jointly affect capacity,
concentration, dependencies and skills. A rule acts upon a system that
reacts.

Legal complexity theory has already challenged reductionist accounts of
the relationship between law and society. Ruhl described the
law--society system as dynamic, adaptive and characterised by emergent
effects, inviting administrative law to recognise non-linearity and
dependencies.\footnote{J. B. RUHL, \emph{Complexity Theory as a Paradigm
  for the Dynamical Law-and-Society System: A Wake-Up Call for Legal
  Reductionism and the Modern Administrative State}, in \emph{Duke Law
  Journal}, 1996, pp.~849 ff.; idem, \emph{Law's Complexity: A Primer},
  in \emph{Georgia State University Law Review}, 2008, pp.~885 ff.} In
Italian scholarship, Pastore linked the rapid development of
technoscience to \emph{law lag} and the need for open, differentiated
and flexible regulation, without excluding the structuring role of
law.\footnote{B. PASTORE, \emph{Tecnologie emergenti, incertezze della
  scienza, regolamentazione giuridica}, in \emph{Teoria e Critica della
  Regolazione Sociale}, 2020, pp.~97 ff., especially pp.~109--112.} This
article builds on these insights but proposes a generative formulation:
not merely to describe the complex ecosystem, but to organise the rules
capable of transforming it.

\emph{Anticipatory governance} is not the same as prediction. Guston
defines it as a distributed capacity to know, reflect upon and intervene
in the implications of science and technology.\footnote{D. H. GUSTON,
  \emph{Understanding ``Anticipatory Governance''}, in \emph{Social
  Studies of Science}, 2014, pp.~218 ff., DOI: 10.1177/0306312713508669.}
The OECD's 2024 framework articulates this capacity through a
value-oriented approach, foresight and technology assessment,
stakeholder engagement, agile and adaptive regulation, and international
cooperation.\footnote{OECD, \emph{Framework for Anticipatory Governance
  of Emerging Technologies}, OECD Science, Technology and Industry
  Policy Papers, no. 165, Paris, 2024, DOI: 10.1787/0248ead5-en.}
Anticipation is thus an institutional practice that prepares for
uncertainty, not a promise of precision. Adaptation likewise has an
established literature. Schrepel connects monitoring and revision with
complexity science while preserving legal certainty and
accountability.\footnote{T. SCHREPEL, \emph{Adaptive Regulation}, in
  \emph{European Journal of Risk Regulation}, 2026, pp.~374 ff., DOI:
  10.1017/err.2026.10087.} Sabel and Zeitlin's \emph{experimentalist
governance} organises framework goals, autonomy, reporting, peer review
and revision.\footnote{C. F. SABEL, J. ZEITLIN, \emph{Experimentalism in
  the EU: Common Ground and Persistent Differences}, in \emph{Regulation
  \& Governance}, 2012, pp.~410 ff., DOI:
  10.1111/j.1748-5991.2012.01157.x; C. F. SABEL, J. ZEITLIN,
  \emph{Experimentalist Governance}, in D. LEVI-FAUR (ed.), \emph{The
  Oxford Handbook of Governance}, Oxford, 2012, DOI:
  10.1093/oxfordhb/9780199560530.013.0012.} Sandboxes and experiments
translate these principles into bounded environments but may undermine
equality, legality and competition if exemptions, access and evaluation
are not regulated.\footnote{S. RANCHORDÁS, \emph{Experimental
  Regulations and Regulatory Sandboxes: Law Without Order?}, in
  \emph{Law and Method}, 2021, DOI: 10.5553/REM/.000064; OECD,
  \emph{Regulatory Experimentation: Moving Ahead on the Agile Regulatory
  Governance Agenda}, OECD Public Governance Policy Papers, no. 47,
  Paris, 2024, DOI: 10.1787/f193910c-en.}

\emph{Robust decision-making} (RDM) shifts the criterion from optimality
to robustness: a strategy is evaluated across many plausible conditions
and against the circumstances in which it fails.\footnote{R. J. LEMPERT,
  S. W. POPPER, S. C. BANKES, \emph{Shaping the Next One Hundred Years:
  New Methods for Quantitative, Long-Term Policy Analysis}, Santa
  Monica, 2003; R. J. LEMPERT, D. G. GROVES, S. W. POPPER, S. C. BANKES,
  \emph{A General, Analytic Method for Generating Robust Strategies and
  Narrative Scenarios}, in \emph{Management Science}, 2006, pp.~514 ff.,
  DOI: 10.1287/mnsc.1050.0472.} \emph{Scenario discovery} identifies
combinations of conditions relevant to a decision's
vulnerability,\footnote{D. G. GROVES, R. J. LEMPERT, \emph{A New
  Analytic Method for Finding Policy-Relevant Scenarios}, in
  \emph{Global Environmental Change}, 2007, pp.~73 ff., DOI:
  10.1016/j.gloenvcha.2006.11.006.} even without a complete formal
model.\footnote{S. W. POPPER, \emph{Robust Decision Making and Scenario
  Discovery in the Absence of Formal Models}, in \emph{Futures \&
  Foresight Science}, 2019, e22, DOI: 10.1002/ffo2.22.}

\emph{Dynamic Adaptive Policy Pathways} (DAPP) are the closest
methodological precursor to the present proposal. They combine initial
actions, alternative pathways, monitoring signals and thresholds
indicating when a measure no longer meets its objectives and must be
supplemented or replaced. They also make path dependence, decision
sequences and the risk of lock-in visible, avoiding commitment to a
single expected future.\footnote{M. HAASNOOT, J. H. KWAKKEL, W. E.
  WALKER, J. TER MAAT, \emph{Dynamic Adaptive Policy Pathways: A Method
  for Crafting Robust Decisions for a Deeply Uncertain World}, in
  \emph{Global Environmental Change}, 2013, vol.~23, no. 2,
  pp.~485--498, DOI: 10.1016/j.gloenvcha.2012.12.006.} The convergence
with the RFAR is substantial in relation to adaptation over time. The
difference concerns the primary analytical unit: DAPP organise sequences
of actions across uncertain futures; the RFAR generates and compares
families of rules, distinguishes transformations from the legal form of
instruments, and constructs contextual equivalences with respect to
specified properties. The two approaches can therefore be integrated:
the RFAR structures the alternatives and their transformations, while
DAPP organise their activation along adaptive pathways.

These theoretical traditions place limits on any claim to originality
for the proposed framework. Feedback, foresight, revision,
experimentation, trajectories and robustness are not inventions of
ruliology. Its distinguishing feature lies not in the isolated presence
of these elements but in their organisation through the relationship
between \textbf{rule space}, \textbf{transformation space},
\textbf{contextual equivalence} and \textbf{trajectories generated by
applying rules}. RDM primarily explores future conditions and strategic
vulnerabilities. Adaptive regulation focuses on the capacity of an
instrument to change over time. Experimentalist governance defines
recursive learning procedures. DAPP organise actions and thresholds
along alternative pathways. The Ruliological Framework for Anticipatory
Regulation (RFAR) proposes a shared framework for representing and
comparing what these approaches explore, sequence and modify.

The analytical differences can be summarised as follows:

\begin{itemize}
\item
  anticipatory governance operates on collective capabilities through
  foresight, engagement and reflexivity, seeking institutional
  preparedness;
\item
  adaptive regulation operates on rules subject to change through
  monitoring and updating, seeking constrained \emph{responsiveness};
\item
  experimentalist governance operates on implementation cycles through
  testing, comparison and revision, seeking recursive learning;
\item
  robust decision-making operates on strategies and vulnerabilities
  through stress tests across many futures, seeking robustness without
  privileging any one future;
\item
  Dynamic Adaptive Policy Pathways operate on sequences of actions,
  signals and thresholds along alternative pathways, seeking to preserve
  options and adaptive capacity;
\item
  the RFAR operates on rules, transformations and trajectories through
  generation, comparison and meta-regulation, seeking a robust
  regulatory portfolio.
\end{itemize}

The proposal must also be distinguished from the tradition of
\emph{regulatory space}. Hancher and Moran used that expression to
analyse how power, resources and regulatory capabilities are distributed
among public and private actors, and how political, legal and cultural
characteristics structure the regulatory arena.\footnote{L. HANCHER, M.
  MORAN, \emph{Organizing Regulatory Space}, in R. BALDWIN, C. SCOTT, C.
  HOOD (eds.), \emph{A Reader on Regulation}, Oxford, 1998, pp.~148 ff.;
  originally in L. HANCHER, M. MORAN (eds.), \emph{Capitalism, Culture
  and Economic Regulation}, Oxford, 1989.} `Rule space' does not merely
rename that theory. \emph{Regulatory space} helps explain who occupies
the arena, who possesses information and who can influence decisions.
Rule space, by contrast, describes the available alternatives and the
transformations they may generate. The two perspectives are
complementary: the distribution of power affects which rules enter the
explorable space and which remain excluded.

\textbf{Ruliology} means the systematic study of possible rules and the
behaviours generated by their execution.\footnote{S. WOLFRAM, \emph{What
  Is Ruliology?}, 12 January 2026, available online on the author's
  website.} Its object is not merely an individual rule in isolation but
the space constituted by rules, the transformations they produce,
equivalences among different rules, and the trajectories that emerge
from their repeated application. The perspective is generative: starting
from a state and a rule, it observes the resulting transformation and
compares possible developments.

The RFAR does not transfer an entire model developed for formal systems
into law. What it draws from ruliology is primarily a grammar of
exploration: defining relevant states, generating families of rules,
observing their effects, comparing trajectories and identifying
constraints that must be preserved. The connection with \emph{generative
social science}\footnote{J. M. EPSTEIN, \emph{Agent-Based Computational
  Models and Generative Social Science}, in \emph{Complexity}, 1999,
  pp.~41 ff., DOI:
  10.1002/(SICI)1099-0526(199905/06)4:5\textless41::AID-CPLX9\textgreater3.0.CO;2-F;
  on graph transformation, see R. HECKEL, \emph{Graph Transformation in
  a Nutshell}, in \emph{Electronic Notes in Theoretical Computer
  Science}, 2006, pp.~187 ff., DOI: 10.1016/j.entcs.2005.12.018.} does
not remove the limits of the analogy: the model selects some
relationships within the ecosystem and does not claim to reproduce its
full institutional and social complexity.

\subsection{4. The Ruliological Framework for Anticipatory
Regulation}\label{the-ruliological-framework-for-anticipatory-regulation}

\subsubsection{4.1 Ecosystem, State and
Rule}\label{ecosystem-state-and-rule}

The starting point is not an isolated legal norm but the
\textbf{regulatory ecosystem}: the relational set of actors,
institutions, infrastructures, resources, technologies and capabilities
contributing to a technology's development, use and governance. The
ecosystem is not simply a `sector'. It encompasses universities,
businesses, agencies, public users, investors, standardisation bodies,
suppliers, skills, components and transnational connections.

The \textbf{regulatory state} \(S_{t}\) is a selective representation of
the ecosystem at a given moment. It is not a complete description of
reality but identifies elements relevant to the decision at hand. The
state may be represented by an attributed graph

\[
G_t=(V_t,E_t,A_t),
\]

where \(V_{t}\) is the set of nodes, \(E_{t}\) is the set of
relationships among nodes, and \(A_{t}\) is the function assigning
attributes to nodes and relationships:

\[
A_t:V_t\cup E_t\longrightarrow\mathcal{A}.
\]

Nodes may represent actors, resources, infrastructures or capabilities;
edges may represent funding, supply, access, cooperation,
standardisation or supervision relationships. Attributes describe
relevant properties such as location, technological maturity, productive
capacity, criticality, substitutability and risk level.

Figure 1 presents an illustrative instance of the regulatory state of
the European quantum supply chain. It does not report the results of an
empirical survey but shows how the components of \(G_{t}\) may be
identified in practice. The selection of nodes, relationships and
attributes must be explicit, justified and open to challenge, since it
determines what the model makes visible and what it leaves in the
background.

\begin{figure}[!htbp]
\centering
\includegraphics[width=1\textwidth,height=\textheight]{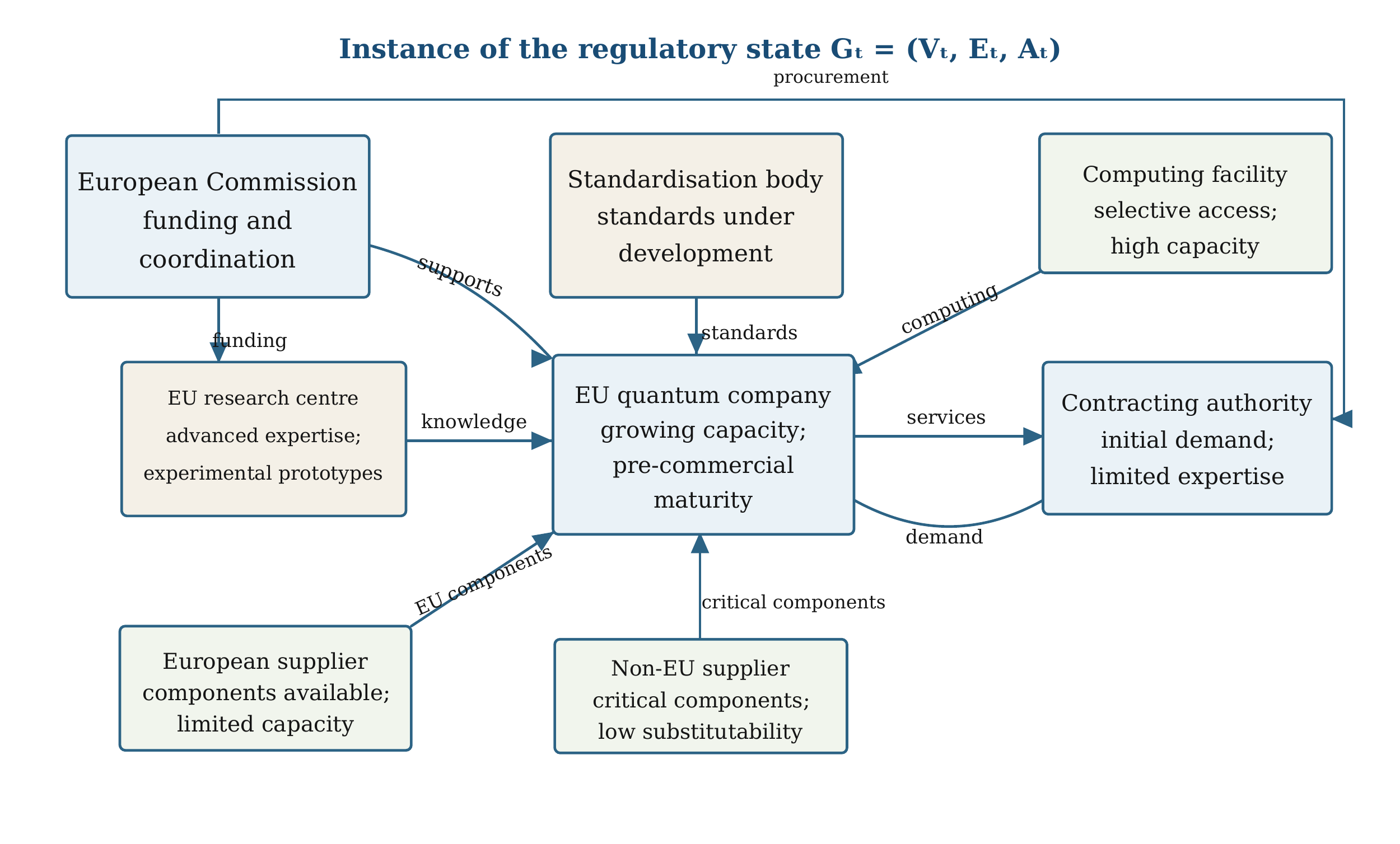}
\caption{Illustrative instance of the regulatory state \(G_t\)
of the European quantum supply chain. Nodes represent actors,
infrastructures and capabilities; edges represent funding, supply,
access, standardisation and public-demand relationships; the indications
within nodes exemplify attributes \(A_t\). The diagram is a conceptual
model, not the outcome of an empirical survey.}
\end{figure}

A \textbf{regulatory rule} \(r\) is a provision or instrument that
enables, prohibits, incentivises or modifies relationships, resources
and attributes. The term includes statutory provisions, standards,
funding criteria, procurement conditions, authorisations, disclosure
obligations and experimental mechanisms. Regulation does not act
directly on technology. It changes which transformations of the
ecosystem are lawful, advantageous, accessible or probable.

A \textbf{regulatory transformation} is the passage between two states
generated by the contextual application of a rule:

\[
T(r,S_t,c):S_t\longrightarrow S_{t+1},
\]

where \(c\) comprises institutional, economic, technological and
geopolitical conditions. Since states are represented by attributed
graphs, the same transformation may be observed as the transition

\[
G_t\longrightarrow G_{t+1}.
\]

The notation does not imply that a rule necessarily determines a single
outcome. It represents the change made possible or probable by the
interaction of rule \(r\), initial state \(S_{t}\), context \(c\) and
actors' responses. The same rule may generate different transformations
in different contexts; different rules may generate similar
transformations with respect to an observed property.

Figure 2 illustrates the application of a procurement rule introducing
requirements for supply diversification and interoperability. In the
initial state, the undertaking depends on a single non-European supplier
of critical components. The rule changes the conditions of access to
public demand and makes entry by an alternative supplier advantageous.
In a possible subsequent state, dependence on the sole supplier is
reduced, and the system offered to the contracting authority must meet
interoperability requirements.

\begin{figure}[!htbp]
\centering
\includegraphics[width=1\textwidth,height=\textheight]{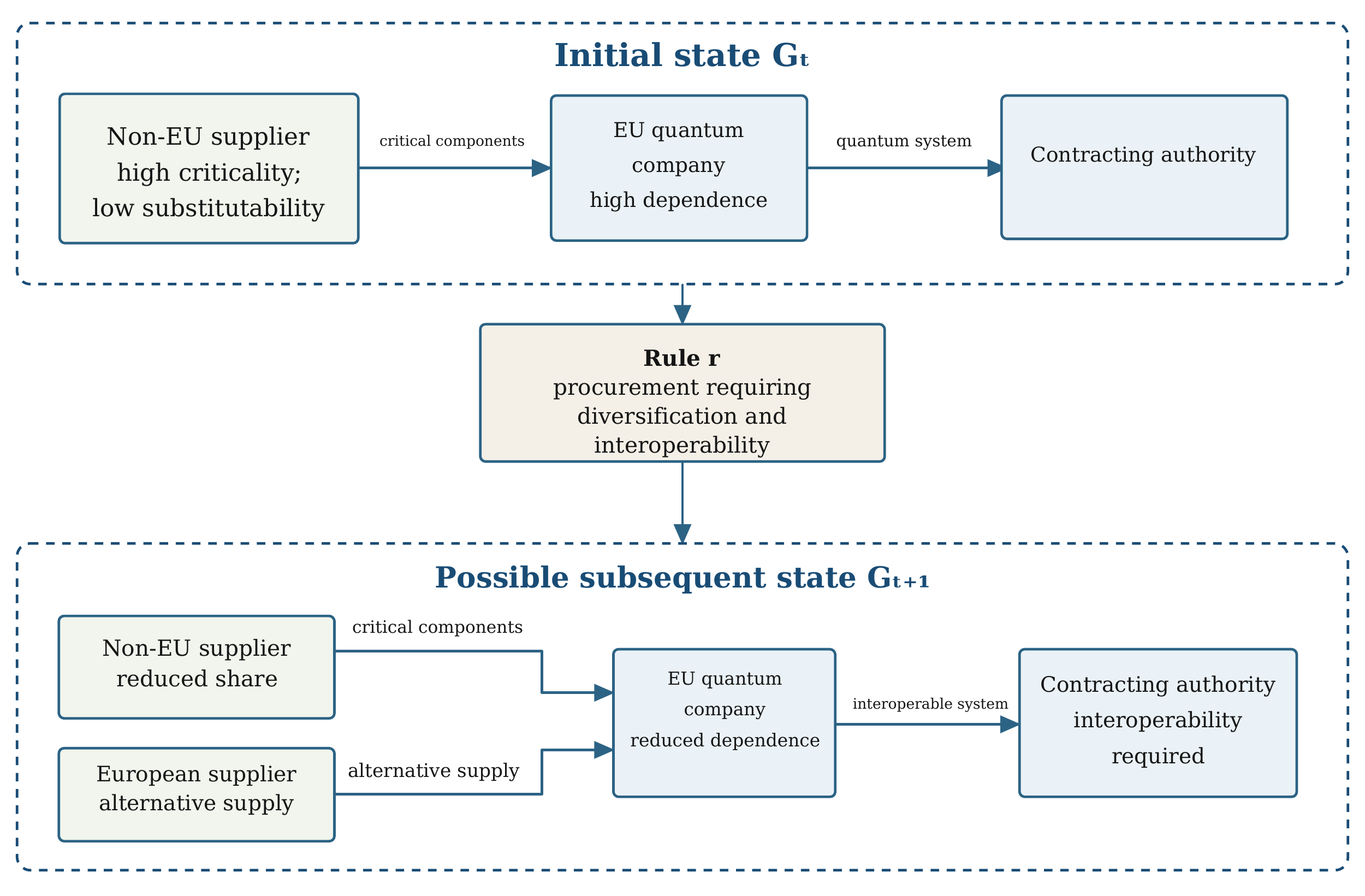}
\caption{Example of a regulatory-state transformation through
a procurement rule. Rule \(r\) introduces diversification and
interoperability requirements. The transition from \(G_t\) to
\(G_{t+1}\) represents a possible transformation, not a necessary
outcome or an empirically demonstrated causal relationship.}
\end{figure}

Transformation is not without costs or side effects. Diversification may
increase supply-chain robustness, but it may also involve initial costs,
adjustment periods, certification difficulties and new coordination
problems. The framework therefore requires a transformation to be
observed against several properties, not merely its principal objective.

\subsubsection{4.2 Rule Space and Contextual
Equivalence}\label{rule-space-and-contextual-equivalence}

The \textbf{rule space} \(\mathcal{R}(S_t,c)\) is the set of
alternatives legally and institutionally applicable in a given state and
context. The \textbf{transformation space} \(\mathcal{T}(S_t,c)\) is the
set of effects achievable through those rules. This distinction prevents
the legal form of an instrument from being confused with what it
produces.

A subsidy, a tax credit and a public contract may all increase
productive capacity. They differ in procedure, competition, risk and
remedies, but they are \textbf{contextually equivalent} if the chosen
evaluation, holding state, context and time horizon constant, assigns
their effects to the same qualitative category, `increase in productive
capacity'. Equivalence here concerns that classification, not
necessarily the magnitude of the increase. The same equivalence does not
extend to concentration, speed or reversibility.

To make this comparison more precise, an evaluation function is
introduced:

\[
\pi_{P,S,c,h}:\mathcal{R}(S,c)\longrightarrow\mathcal{Y}_P,
\]

which associates each rule with the profile of transformations relevant
to property \(P\), starting from state \(S\), in context \(c\) and over
horizon \(h\). In this notation, \(S\) denotes the initial state
generically, which may be identified with \(S_t\) at the time
considered, while \(\mathcal{Y}_P\) is the space of possible
transformation profiles associated with property \(P\). We therefore
define:

\[
r_1\sim_{P,S,c,h}r_2
\quad\Longleftrightarrow\quad
\pi_{P,S,c,h}(r_1)=\pi_{P,S,c,h}(r_2).
\]

The subscripts denote:

\begin{itemize}
\item
  \(P\): the regulatory property with respect to which the comparison is
  made, such as productive capacity, accessibility, concentration or
  security;
\item
  \(S\): the initial state of the ecosystem;
\item
  \(c\): the institutional, economic, technological and geopolitical
  context;
\item
  \(h\): the time horizon or number of steps considered.
\end{itemize}

Thus defined, \(\sim_{P,S,c,h}\) is an equivalence relation, because it
is reflexive, symmetric and transitive. This does not mean that
\(r_{1}\) and \(r_{2}\) are identical or interchangeable legal norms in
every respect. It means only that the two rules belong to the same class
with respect to the property and conditions specified by the subscripts.

In empirical applications, it is often possible to establish only that
effects fall within a declared tolerance. In that case it is preferable
to write

\[
r_1\approx_{P,S,c,h,\varepsilon}r_2,
\]

where \(\varepsilon\) is the permitted deviation threshold. This is a
contextual proximity relation and not necessarily an equivalence
relation in the mathematical sense, because transitivity need not hold.

\begin{figure}[!htbp]
\centering
\includegraphics[width=1\textwidth,height=\textheight]{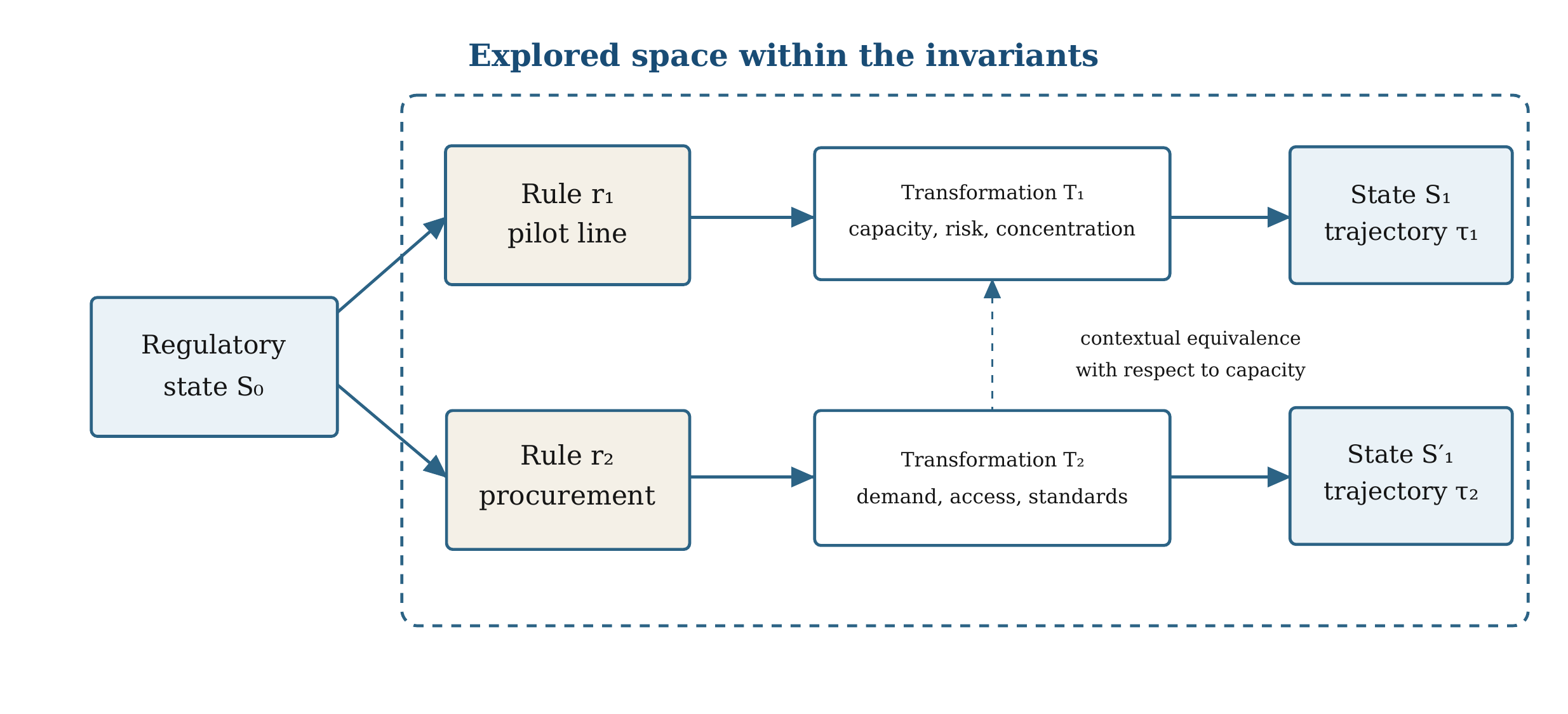}
\caption{Contextual equivalence between different rules.
Starting from the same regulatory state \(S_0\), rule \(r_1\),
concerning a pilot line, and rule \(r_2\), concerning procurement,
generate different transformations. The transformations may be regarded
as equivalent with respect to the property `capacity', while diverging
in terms of risk, concentration, demand, access and standardisation. The
diagram represents a conceptual model, not the result of a simulation.}
\end{figure}

Figure 3 illustrates the necessarily parameter-dependent character of
equivalence. A pilot line and a procurement programme may be regarded as
equivalent with respect to \(P = \text{capacity}\) if, holding the
state, context and time horizon constant, their effects are assigned to
the same qualitative category, `increase in ecosystem capacity'. This
equivalence does not imply that the two instruments produce
quantitatively identical increases. Nor do they belong to the same class
with respect to other properties: the pilot line primarily affects the
availability of infrastructure and the reduction of technological risk,
while procurement affects demand, market access and the diffusion of
standards. Equivalence with respect to one property therefore cannot
automatically be extended to the entire regulatory transformation.

\textbf{Transformation cost} describes the burdens associated with the
transition. It is not a mathematical distance. It may include financial
cost, time, administrative burden, concentration of power, loss of
future options, difficulty of reversal, distributional effects and risks
to rights. These dimensions must not automatically be aggregated into a
single value, because aggregation would conceal political and
constitutional choices.

\subsubsection{4.3 Trajectories, Invariants and
Robustness}\label{trajectories-invariants-and-robustness}

The repeated application of rules, together with actors' responses and
changes in context, generates a \textbf{regulatory trajectory}

\[\tau = (S_{0},S_{1},\ldots,S_{n}).\]

A measure adopted in state \(S_{0}\) may attract businesses, change the
distribution of scientific and technological skills, consolidate a
standard and create new dependencies. These changes alter the subsequent
state and condition the effectiveness of rules applicable at later
stages. The trajectory thus represents a path-dependent institutional
evolution, not a purely technical sequence or a deterministic
prediction.

\textbf{Invariants} delimit the set of admissible trajectories. The term
is used in a normative sense: it does not mean that certain empirical
quantities must remain numerically constant, but that certain
constraints cannot be sacrificed during exploration. Fundamental rights,
non-discrimination, proportionality, transparency, controllability and
the availability of remedies are not outcomes to be maximised alongside
others. They are conditions of admissibility. An economically effective
trajectory is not acceptable from a regulatory standpoint if it depends
on discrimination, disproportionate surveillance, uncontrollable
concentration or restrictions on judicial protection.

\textbf{Control thresholds} must be distinguished from invariants. They
translate objectives and risks into observable conditions: a maximum
level of dependence on a single supplier, a minimum notice period, a
proportion of available capacity or a concentration indicator.
Thresholds may be reconsidered through reasoned procedures; invariants,
by contrast, operate as limits on the selection of trajectories.

\textbf{Regulatory robustness} is the capacity of a rule or regulatory
portfolio to maintain acceptable outcomes and respect invariants across
a plurality of plausible trajectories, while preserving opportunities
for correction and learning. It is not identical to adaptability, which
concerns whether a rule can be modified; monitoring, which concerns the
production of knowledge; or resilience, which concerns the ecosystem's
ability to absorb disturbances. A regulatory regime may be easily
modifiable but insufficiently robust, or robust in the face of a shock
yet incapable of learning from unexpected transformations.

\subsubsection{4.4 A Minimal Operational
Procedure}\label{a-minimal-operational-procedure}

The RFAR can be applied through seven interconnected steps forming a
cycle:

\begin{enumerate}
\def\labelenumi{\arabic{enumi}.}
\item
  \textbf{Observe and represent the state.} Identify relevant actors,
  capabilities, relationships, dependencies and risks, specifying what
  the representation excludes and the reliability of the information
  available.
\item
  \textbf{Generate families of rules.} Construct alternatives from
  different categories---incentives, prohibitions, standards,
  procurement, disclosure obligations or experimentation---without
  prematurely selecting a single instrument.
\item
  \textbf{Define criteria and filter alternatives.} Specify the
  objectives pursued, the invariants delimiting legally admissible
  transformations, and the thresholds indicating when a trajectory
  requires attention, revision or termination. Exclude alternatives
  incompatible with the invariants.
\item
  \textbf{Describe and compare transformations.} For each family, assess
  direct effects, foreseeable responses by actors, interactions with
  other rules and second-order effects. The comparison must cover
  multiple properties rather than the principal objective alone.
\item
  \textbf{Construct and select a portfolio.} Consider rules in
  combination as well as individually. Selection favours portfolios
  capable of maintaining acceptable outcomes across different
  trajectories, avoiding excessive dependence on a single prediction.
\item
  \textbf{Monitor thresholds and trajectories.} Monitoring checks
  whether the observed state departs from expected transformations,
  whether control thresholds are crossed or whether effects incompatible
  with invariants emerge.
\item
  \textbf{Revise, replace or terminate.} Meta-rules determine who may
  change the regime, on the basis of which data, with what
  justification, within which time limits, and subject to which forms of
  oversight and accountability. The outcomes of revision feed into a new
  representation of the state and reopen the cycle.
\end{enumerate}

The framework fails if it cannot distinguish alternatives that existing
approaches would treat in the same way; if it does not make criteria for
revision, termination and responsibility explicit; if the cost of
constructing the space exceeds the decision-relevant information
obtained; or if formalism conceals the distributional and constitutional
choices it is intended to make visible.

\subsection{5. Governing the European Quantum Supply
Chain}\label{governing-the-european-quantum-supply-chain}

The \emph{Quantum Europe Strategy: Quantum Europe in a Changing World},
adopted by the Commission on 2 July 2025, sets out five areas: research
and innovation, infrastructures, strengthening the ecosystem, space and
dual-use, and skills.\footnote{European Commission Communication of 2
  July 2025, \emph{Quantum Europe Strategy: Quantum Europe in a Changing
  World}, COM(2025)363 final, pp.~3--4.} The document connects them
through an iterative cycle and a governance framework. In particular, it
envisages continuous iteration between discovery, development, testing
and deployment, supported by testbeds, pilot lines, KPIs, milestones,
benchmarking and feedback loops.\footnote{Ibid., pp.~19--21.} The
Strategy itself describes its approach as an `evolving framework', an
undertaking in progress.\footnote{Ibid., p.~22.}

The European cycle already includes monitoring and feedback; what
remains less explicit is how heterogeneous alternatives are generated,
compared and replaced.

The supply chain is the principal test case. The Strategy recognises
vulnerabilities involving rare materials, precision components, control
electronics and software. It envisages Union-level assessments and
identifies four families of responses: supplier diversification,
European manufacturing capacity, partnerships with supplier countries,
and risk-sharing mechanisms.\footnote{Ibid., pp.~15--16.} The JRC has
also documented the coexistence of scientific excellence, fragmented
investment, differing levels of maturity and the limited conversion of
research capacity into patents and industrial scale.\footnote{A. M.
  LEWIS ET AL., \emph{Future Directions for Quantum Technology in
  Europe: An Analysis of Policy Questions}, JRC141050, Luxembourg, 2025,
  DOI: 10.2760/0891087; OECD, \emph{A Quantum Technologies Policy
  Primer}, OECD Science, Technology and Industry Policy Papers, no. 371,
  Paris, 2025, DOI: 10.1787/fd1153c3-en.}

A possible information layer for constructing the regulatory state is
provided by the \emph{Quantum Criticality Index} (QCI), which assesses
quantum materials and components along three axes: supply risk,
substitutability and strategic relevance.\footnote{D. CHO, M. KOP, M.-H.
  LEE, \emph{Strategic Governance of Quantum Supply Chains: A
  Criticality-Based Framework for Risk, Resilience, and Data-Driven
  Foresight}, in \emph{EPJ Quantum Technology}, 2026, vol.~13, art. 43,
  DOI: 10.1140/epjqt/s40507-026-00486-y.} In the language of the RFAR,
these dimensions may be represented as attributes \(A_{t}\) of the nodes
and relationships of \(G_{t}\). Indicators of manufacturing
concentration, dependence on imports, stability of supplier countries,
availability of substitutes and dual-use relevance may also contribute
to defining control thresholds. The QCI does not replace rule space: it
provides a diagnosis of criticality and useful monitoring signals, while
the RFAR compares the transformations that can be produced through
procurement, capacity support, diversification, partnerships or
standards. The same criticality may indeed warrant different
alternatives, which must also be assessed in terms of distributional
effects, reversibility, competition and compatibility with invariants.

A \emph{prediction-first} question would be: which component will become
critical, when will dependence arise, and what capacity will be needed?
Foresight remains useful, but its answer may be fragile when
technological timelines, demand and geopolitics evolve together. The
RFAR instead poses four operational questions: which rules alter
dependence; which transformations do they share; where do they diverge;
and which signals should trigger revision or termination?

Consider four families of rules.

\textbf{Pilot lines and supply-side support.} Funding, public equity or
risk-sharing may reduce investment risk and increase domestic capacity.
The desired transformation is the transition from laboratory to
production. Side effects include beneficiary selection, concentration,
dependence on subsidies and technological lock-in.

\textbf{Coordinated procurement.} Predictable public demand may support
investment and impose requirements concerning interoperability, security
or origin. The local transformation may be equivalent to that of
supply-side support with respect to productive capacity. It nevertheless
differs in the distribution of risk, market access, compatibility with
competition and international commitments.

\textbf{Diversification and partnerships.} Agreements with trusted
suppliers may reduce concentration and lead times without bringing all
production in-house. The trajectory preserves openness and
specialisation but remains exposed to political dependencies and
reciprocity clauses.

\textbf{Modular standards.} Common requirements may improve
interoperability, substitutability and supply-chain transparency. If
adopted too early, they may entrench an immature architecture. Research
on European quantum standardisation emphasises precisely the
relationship between maturity, coordination and the need to develop
standards without obstructing innovation.\footnote{O. VAN DEVENTER ET
  AL., \emph{Towards European Standards for Quantum Technologies}, in
  \emph{EPJ Quantum Technology}, 2022, art. 33, DOI:
  10.1140/epjqt/s40507-022-00150-1.}

Comparison yields a portfolio. Interactions matter: proprietary
requirements may undermine modularity; concentrated support may reduce
diversification; and partnerships may transfer dependencies towards
non-substitutable components.

Meta-rules must make this exploration subject to legal oversight. One
example is an annual review based on publicly available thresholds for
supplier concentration, productive capacity, replacement time,
technological delay and cost. Crossing a threshold does not
automatically determine which measure is adopted. It triggers a duty to
reconsider the portfolio, compare alternatives, justify the choice and
verify proportionality and compatibility with invariants. The authority
must also publish aggregated data and permit scrutiny by excluded
operators and judicial bodies.

As at 30 August 2026, EUR-Lex records an initiative for a regulation
entitled the EU Quantum Act. Preparatory document Ares(2025)9260984 sets
out an indicative timetable for the second half of 2026. The
consultation was published on 29 October 2025 and extended until 15
December 2025; no Commission proposal (COM) for the Quantum Act was yet
available in the official sources consulted.\footnote{European
  Commission, \emph{Commission Invites Contributions to Shape Future EU
  Quantum Act}, press release of 31 October 2025, updated on 24 November
  2025; document Ares(2025)9260984, \emph{EU Quantum Act --- Call for
  Evidence for an Impact Assessment}. Source status verified as at 30
  August 2026.} The considerations offered here are therefore \emph{de
iure condendo}. They suggest that the future act should not merely list
instruments and competences but also establish the procedure governing
their comparison and updating.

\subsection{6. Three Stress Tests: Openness and Security, Standards,
Human
Capital}\label{three-stress-tests-openness-and-security-standards-human-capital}

\subsubsection{6.1 Scientific Openness, Autonomy and
Security}\label{scientific-openness-autonomy-and-security}

The European Strategy seeks to protect strategic interests while
maintaining openness towards trusted partners. It envisages joint
programmes, reciprocal access to infrastructures, alignment of
intellectual property arrangements and cooperation on standards,
alongside investment and export controls and protection for dual-use
technologies.\footnote{Communication COM(2025)363 final, cited above,
  pp.~20--22.} The choice cannot be reduced to an
openness-versus-closure dichotomy.

Open access, selective cooperation, screening and restrictions alter
flows of knowledge, capital and talent. They may reduce the same
vulnerability but differ in their effects on research freedom,
attractiveness and reciprocity. The resulting portfolio is
differentiated by component and maturity, with restrictions that are
narrowly tailored, justified and reviewable.

The invariant is twofold. Security cannot become an indeterminate
formula that removes decisions from scrutiny. Openness cannot prevent
the protection of essential capabilities. The meta-rule must specify the
competent authority, evidence, duration, review, scientific consultation
and remedies.

\subsubsection{6.2 Standardisation and
Certification}\label{standardisation-and-certification}

Standards coordinate immature markets and enable interoperability, but
they may embed premature choices. Rule space encompasses voluntary
standards, procurement, certification, testing and disclosure
obligations.

The proposal for \emph{standards-first} quantum governance regards
standards as a flexible and internationally harmonisable means of
governing technologies that are still immature, while preserving
innovation, security and cooperation.\footnote{M. ABOY, U. GASSER, I. G.
  COHEN, M. KOP, \emph{Quantum Technology Governance: A Standards-First
  Approach}, in \emph{Science}, 2025, vol.~389, no. 6760, pp.~575--578,
  DOI: 10.1126/science.adw0018.} This approach strengthens the
centrality of standardisation but does not entail a
\emph{standards-only} strategy. Within the RFAR, standards remain a
family of rules to be compared with procurement, testbeds, certification
and disclosure obligations. The added value of comparison lies in
determining whether different instruments produce locally equivalent
properties and where they diverge in terms of market access, compliance
costs, concentration, security and reversibility.

Two rules may improve comparability without being equivalent in terms of
access for SMEs. Certification increases trust but creates fixed costs;
a testbed produces information with less rigidity.

A meta-rule may link the binding force of a standard to technological
maturity. In early stages, it favours measurement methods, vocabularies
and modular interfaces; as use and risk increase, it may activate more
stringent requirements. Sunset and review clauses prevent a standard
initially adopted as a bridge from turning into unexamined lock-in.

\subsubsection{6.3 Human Capital}\label{human-capital}

The Strategy recognises a strong academic base and shortages of applied
skills, including software engineering, system integration and quantum
cybersecurity. It envisages a European Quantum Skills Academy, mobility
programmes, apprenticeships and researchers-in-residence, together with
continuous monitoring of skills demand.\footnote{European Commission,
  \emph{Quantum Europe Strategy: Quantum Europe in a Changing World},
  COM(2025)363 final, cited above, pp.~17--20.}

An academy coordinates curricula; a fellowship changes mobility; an
apprenticeship links training with industry; a researcher-in-residence
transfers knowledge. The `number of people trained' is not enough:
distribution, retention, diversity and applied capabilities also matter.

The RFAR suggests avoiding policies tied to precise forecasts of future
occupations. A modular portfolio, accompanied by meta-rules that update
curricula and incentives on the basis of verifiable signals, is
preferable. Here too, invariants exist: non-discriminatory access,
transparent selection procedures, academic freedom and protection of
data used for monitoring.

\subsection{7. Meta-rules, Safeguards and
Limits}\label{meta-rules-safeguards-and-limits}

Exploration is not synonymous with unlimited discretion. A sandbox
without criteria may turn uncertainty into privilege. A review clause
without thresholds may legitimise perpetual instability. Governance
entrusted to experts may move distributional choices beyond the
democratic process. The decisive transition from vocabulary to
regulation therefore lies in \textbf{meta-rules}.

Meta-rules cover competence, admissibility, information, thresholds,
justification, reversibility and oversight. They identify who makes
selections, which trajectories are excluded, what data are required,
when to correct or terminate a measure, and how participation and
avenues of appeal are guaranteed.

\begin{figure}[!htbp]
\centering
\includegraphics[width=1\textwidth,height=\textheight]{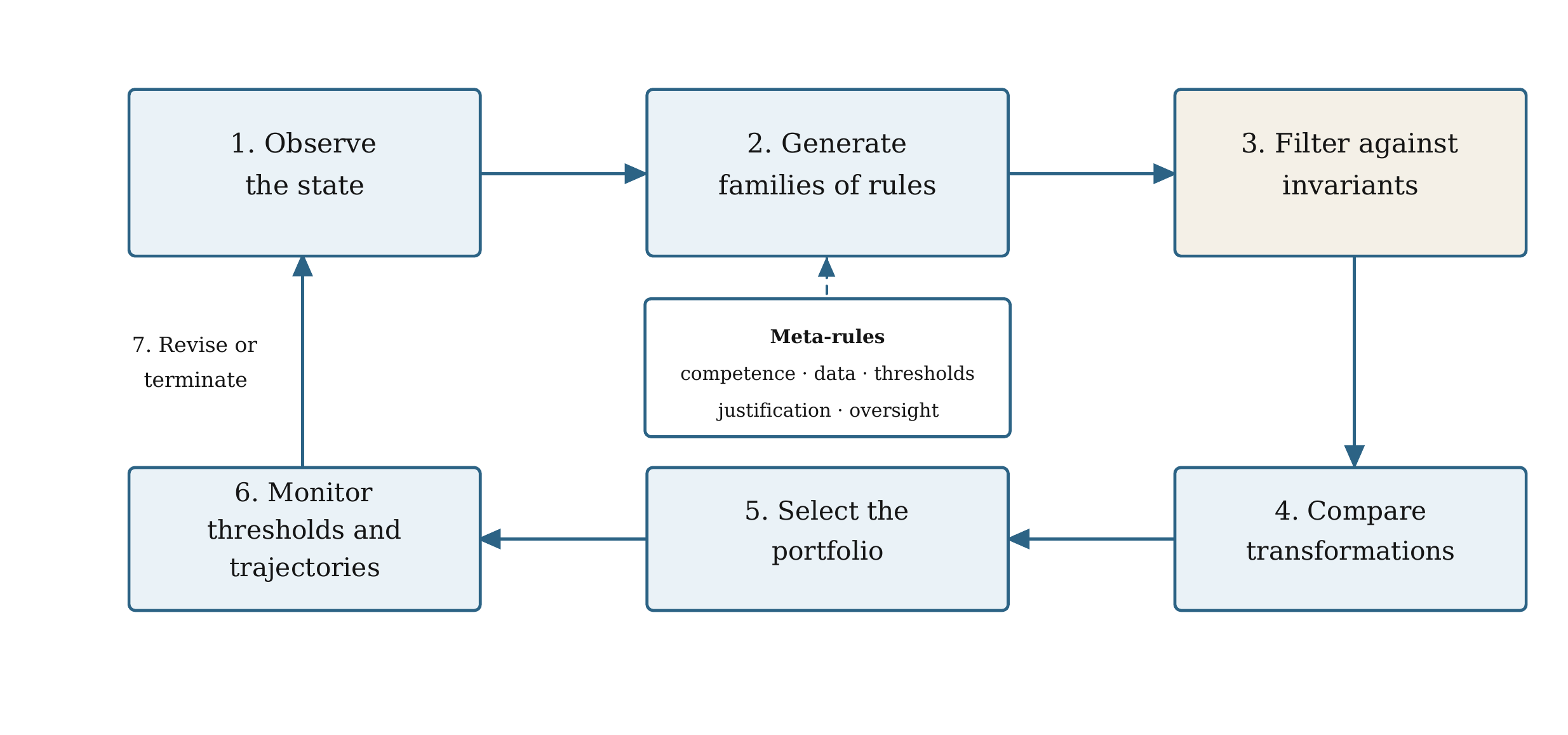}
\caption{The RFAR institutional cycle: observation,
generation, filtering through invariants, comparison, selection,
monitoring, and revision or termination.}
\end{figure}

The RFAR does not replace precaution. It specifies how to compare
responses and update decisions without sacrificing invariants; it does
not authorise experiments that infringe inalienable rights.

Four limitations remain. The first is \textbf{representational}: every
graph omits relationships and may reflect the perspective of the
authority constructing it. The second is \textbf{causal}: an observed
transformation does not, by itself, prove that the rule caused it. The
third is \textbf{institutional}: exploring alternatives requires data,
expertise and coordination, whose costs may exceed the informational
benefit. The fourth is \textbf{democratic}: frequent updating may weaken
political accountability unless objectives and thresholds are
established publicly.

The framework must therefore remain falsifiable. Future validation
should compare decision-making cases conducted with and without the
RFAR, assessing whether it broadens the alternatives considered, reveals
overlooked vulnerabilities, improves the justification of revisions and
reduces lock-in without imposing disproportionate burdens. Until then,
the RFAR remains a methodological proposal, not a regulatory technology
of demonstrated effectiveness.

\subsection{8. Conclusions}\label{conclusions}

Governing emerging technologies is particularly complex because of both
the speed of the transformations they generate and the resistance of
major lobbying interests to regulatory models. Nevertheless, rules are
essential to public decision-making and to evaluating the impact those
decisions can produce over time. From this perspective, the public
interest must remain central and must not evaporate into the algorithmic
ocean.

In this connection, the European and Italian regulatory framework on
artificial intelligence, comprising Regulation (EU) 2024/1689, the
so-called \emph{AI Act}, and Italian Law No.~132 of 23 September
2025,\footnote{Regulation (EU) 2024/1689 of the European Parliament and
  of the Council of 13 June 2024 laying down harmonised rules on
  artificial intelligence, in the \emph{Official Journal of the European
  Union}, L, 12 July 2024; Italian Law No.~132 of 23 September 2025,
  \emph{Disposizioni e deleghe al Governo in materia di intelligenza
  artificiale}, in the \emph{Gazzetta Ufficiale}, General Series, no.
  223, 25 September 2025.} is a model of considerable interest,
distinguished from other international approaches by the attention it
pays to fundamental rights and, more generally, to the human
person.\footnote{See F. Gallo, \emph{Cyberspazio, diritti e tutele}, in
  A. F. Uricchio, C. Caldarola (eds.), \emph{L'intelligenza artificiale
  tra regolazione ed esperienze applicative. AI Act (regolamento UE) e
  disegno di legge governativo}, Cacucci, Bari, 2025, pp.~15--29,
  especially p.~16. The author emphasises the need for European
  institutions to define a clear regulatory framework for artificial
  intelligence. He observes that while, on the one hand, we now live in
  something resembling a metaverse populated by `intelligent' objects
  that speak to, listen to and assist us in our many everyday
  activities, on the other hand, concerns are growing about the
  potential dark sides of digitalisation, the `artificial' character of
  its products, the risks of manipulation, the absence of adequate
  oversight and the possible reduction in employment it may cause. Hence
  the need for European legislation capable of preventing such effects
  over time.} Moreover, while the Trump administration's \emph{Action
Plan}\footnote{See the White House, \emph{Winning the Race: America's AI
  Action Plan}, Washington, 23 July 2025, which endorses the so-called
  unleashing of artificial intelligence through the removal of
  regulatory barriers and the absolute primacy of the market. Executive
  Order 14148 of 20 January 2025 revoked Executive Order 14110 of the
  Biden administration; the Office of Management and Budget also removed
  many earlier constraints. The Plan strengthens the \emph{National AI
  Research Resource} and provides for the preparation of a new national
  strategic research-development plan. This approach is not new: already
  in the 1990s, at the dawn of the spread of the Internet,
  self-regulation was preferably left to the market, affirming the
  principle of online freedom and the substantive absence of rules. The
  result was the granting of excessive discretion to major platforms,
  with a corresponding weakening of trust in legislators. On this
  subject see J. R. Reidenberg, \emph{Lex Informatica: The Formulation
  of Information Policy Rules Through Technology}, in \emph{Texas Law
  Review}, 1998, vol.~76, no. 3, pp.~553--593, who points out that
  `technological capabilities and system-design choices impose rules on
  participants'. This approach was subsequently taken up and
  systematised by L. Lessig, \emph{Code Is Law: On Liberty in
  Cyberspace}, in \emph{Harvard Magazine}, vol.~102, no. 3,
  January--February 2000, pp.~37 ff., according to whom computer code
  constitutes a form of `law', as it can affect individual freedoms and
  shape the exercise of rights in digital space.} removes obstacles and
retreats from regulation, consistently with the traditional reluctance
to impose rigid forms of state control, China adopts a strongly
centralised model based on pervasive and often intrusive oversight by
state authorities, as evidenced by the \emph{Provisions on the
Administration of Deep Synthesis Internet Information Services} of 2022
and the \emph{Interim Measures for the Management of Generative
Artificial Intelligence Services} of 2023. Despite its distinctive
characteristics, the Russian Federation displays a similar tendency,
whereby technological regulation is intertwined with strategies of
totalitarian control, without foregoing systematic and coordinated
cyberattacks against the digital platforms and infrastructures of
countries regarded as hostile or competing.\footnote{For the Chinese
  instruments mentioned, see the \emph{Provisions on the Administration
  of Deep Synthesis Internet Information Services}, issued in 2022 and
  entering into force on 10 January 2023, and the \emph{Interim Measures
  for the Management of Generative Artificial Intelligence Services},
  issued in 2023 and entering into force on 15 August 2023. For hybrid
  and cyber activities attributed to the Russian Federation, see Council
  of the European Union, \emph{Russian Hybrid Threats: EU Sanctions},
  official source consulted on 30 August 2026. The political and
  evaluative characterisation of the Russian model represents the
  authors' argument.} Contrasting paradigms thus emerge: on the one
hand, deregulated market fundamentalism, which entrusts the governance
of strategic sectors to economic dynamics; on the other, all-pervasive
state control, which risks becoming a suffocating web capable of
tightening its grip on the individual. Although grounded in different
premises, both models reveal problematic consequences for individual and
collective safeguards.

The European model, by contrast, occupies an intermediate and innovative
position, aspiring to balance freedom and security, autonomy and
responsibility, almost as a `third pole' capable of reconciling equity
and transparency, markets and dignity, technology and law. It places
fundamental rights at the centre of regulation and adopts a risk-based
approach as its organising principle, seeking to balance the promotion
of innovation with the protection of the individual. The path is not
without obstacles, including risks that particularly intrusive
technological applications may undermine fundamental rights.
Nevertheless, the solution adopted by the European institutions
following a process of public consultation is commendable and worthy of
recognition, as it is inspired by shared values and a far-sighted
conception of law, oriented towards the future and capable of ensuring a
full balance between technological innovation and fundamental
rights.\footnote{As D.-U. Galetta and B. Cappiello emphasise in \emph{Il
  Regolamento UE sull'intelligenza artificiale e le amministrazioni
  pubbliche: basi giuridiche, obiettivi e limiti di un esperimento
  normativo dagli esiti incerti}, in A. F. Uricchio, C. Caldarola
  (eds.), \emph{L'intelligenza artificiale tra regolazione ed esperienze
  applicative. AI Act (regolamento UE) e disegno di legge governativo},
  Cacucci, Bari, 2025, pp.~59--86, the exercise of the Union's
  legislative powers in AI is justified, on the one hand, by the need to
  approximate sector-specific laws under Article 114 TFEU in order to
  promote internal-market integration and, on the other, by the aim of
  protecting natural persons in specific cases with regard to
  personal-data processing under Article 16 TFEU. The authors also
  emphasise that the AI Act's primary objective is to `improve the
  functioning of the internal market by establishing a uniform legal
  framework, particularly with regard to the development, marketing and
  use of artificial intelligence in accordance with Union values'. On
  this subject see also S. Poli, \emph{Il rafforzamento della sovranità
  tecnologica europea e il problema delle basi giuridiche}, in \emph{I
  Post di AISDUE}, III (2021), `Atti convegni AISDUE' section, no. 5, 20
  December 2021, pp.~69 ff.; M. Inglese, \emph{Il regolamento
  sull'intelligenza artificiale come atto per il completamento e il buon
  funzionamento del mercato interno?}, in \emph{Quaderni AISDUE},
  special issue no. 2/2024, pp.~1 ff.}

The constructive relationship between public policies and the right to
the future must clearly remain central, also through transparent and
informed decision-making. This goes beyond the capacity to anticipate
the effects of public choices. The literature on anticipation,
adaptation, experimentation and robustness has already established as
much. The present article proposes a regulatory model that is not
limited to producing rules but is also capable of assessing their
impact, transformations and trajectories.

The distinction between rule space and transformation space makes
different legal instruments comparable without conflating them.
Contextual equivalence shows that two rules may produce the same effect
while diverging in respect of other values. Invariants prevent
exploration from sacrificing rights and safeguards. Meta-rules assign
competences and impose thresholds, justification, revision and
termination. Robustness replaces optimality with respect to a privileged
future with acceptable performance across multiple trajectories.

The European Quantum Strategy already contains many of the necessary
elements: an iterative cycle, testbeds, KPIs, benchmarking, monitoring
and feedback. The future Quantum Act offers an opportunity to regulate
not only the instruments but also the procedure that generates, compares
and corrects them.

Governing the rule space does not mean predicting every outcome or
automatically executing law. It means developing greater awareness,
making public decisions rational, informed and guided by the aim of
producing the common good, both present and future, rather than merely
being effective.

\subsection{Biographical Notes}\label{biographical-notes}

Francesco Scagliola is Professor of Electroacoustic Music Composition at
the ``Niccolò Piccinni'' Conservatoire in Bari, where he coordinates the
School of Electronic Music and the PhD programme in Computational
Electroacoustic Music Composition, Generative Algorithmic Models and
Computational Complexity.

Antonio Felice Uricchio is a former Rector of the University of Bari
Aldo Moro and former President of ANVUR; he is currently a member of the
Board of AVEPRO (Holy See).

\subsection{Final Declarations}\label{final-declarations}

\textbf{Availability of materials.} The article uses only publicly
available academic sources and institutional documents; no research
dataset was created.

\textbf{Ethics.} The research is theoretical and documentary; it does
not involve human participants or personal data. Dual-use issues are
addressed at the governance level, without operational technical
information.

\textbf{Conflicts of interest.} The authors declare no conflicts of
interest relating to the content of the article.

\textbf{Funding.} The research received no specific funding.

\textbf{Author contributions.} Both authors contributed to the
conception, analysis and revision of the article. Francesco Scagliola
was particularly responsible for the computational formalism; Antonio
Felice Uricchio for the legal and institutional framing.

\textbf{Use of artificial intelligence systems.} Generative artificial
intelligence tools were used under the authors' supervision for source
checking and editorial support. The authors remain responsible for the
content, citations and final wording.

\subsection{Bibliography}\label{bibliography}

\begin{itemize}
\item
  ABOY M., GASSER U., COHEN I. G., KOP M., \emph{Quantum Technology
  Governance: A Standards-First Approach}, in \emph{Science}, 2025,
  vol.~389, no. 6760, pp.~575-578, DOI: 10.1126/science.adw0018.
\item
  ABRESCIA M., \emph{Un diritto al futuro: analisi economica del
  diritto, Costituzione e responsabilità tra generazioni}, in BIFULCO
  R., D'ALOIA A. (eds.), \emph{Un diritto per il futuro. Teorie e
  modelli dello sviluppo sostenibile e della responsabilità
  intergenerazionale}, Jovene, Naples, 2008, pp.~161 ff.
\item
  ALPA G., \emph{Prefazione}, in RUFFOLO U. (ed.), \emph{Intelligenza
  artificiale. Il diritto, i diritti, l'etica}, Giuffrè Francis
  Lefebvre, Milan, 2020, pp.~XVII ff.
\item
  ARCONZO G., \emph{La sostenibilità delle prestazioni previdenziali e
  la prospettiva della solidarietà intergenerazionale. Al crocevia tra
  gli artt. 38, 81 e 97 Cost.}, in \emph{Osservatorio costituzionale
  AIC}, issue 3/2018, pp.~627 ff.
\item
  BACON F., \emph{Meditationes Sacrae} (1597), in SPEDDING J., ELLIS R.
  L., HEATH D. D. (eds.), \emph{The Works of Francis Bacon}, London,
  1857-1874.
\item
  BIFULCO R., D'ALOIA A. (eds.), \emph{Un diritto per il futuro. Teorie
  e modelli dello sviluppo sostenibile e della responsabilità
  intergenerazionale}, Jovene, Naples, 2008.
\item
  BOCCIA F., LEONARDI R. (eds.), \emph{The Challenge of the Digital
  Economy: Markets, Taxation and Appropriate Economic Models}, Palgrave
  Macmillan, Cham, 2016, DOI: 10.1007/978-3-319-43690-6.
\item
  CATERINI E., \emph{La «sostenibilità» nell'esperienza giuridica
  contemporanea. In occasione della discussione svoltasi nell'Ateneo
  barese}, in BENÍTEZ D. A., FAVA C. (eds.), \emph{Sostenibilità: sfida
  o presupposto?}, Milan, 2019, pp.~77 ff.
\item
  CATERINI E., \emph{Sostenibilità e ordinamento civile. Per una
  riproposizione della questione sociale}, ESI, Naples, 2018.
\item
  CHO D., KOP M., LEE M.-H., \emph{Strategic Governance of Quantum
  Supply Chains: A Criticality-Based Framework for Risk, Resilience, and
  Data-Driven Foresight}, in \emph{EPJ Quantum Technology}, 2026,
  vol.~13, art. 43, DOI: 10.1140/epjqt/s40507-026-00486-y.
\item
  CONTALDO A., DAINOTTI F., \emph{Diritto e tecnologie delle reti di
  informazione}, Edizioni Giuridiche Simone, Naples, 2005.
\item
  COTTLE T. J., KLINEBERG S. L., \emph{The Present of Things Future:
  Explorations of Time in Human Experience}, The Free Press, New York,
  1974.
\item
  DE TONI A. F., \emph{Prefazione}, in ADINOLFI P., URICCHIO A. F.
  (eds.), \emph{Valutare la complessità. La performance nelle
  organizzazioni pubbliche}, Carocci, Rome, 2025, p.~15.
\item
  DRAETTA U., \emph{Internet e commercio elettronico nel diritto
  internazionale dei privati}, Giuffrè, Milan, 2001.
\item
  EINAUDI L., \emph{Conoscere per deliberare}, first instalment of
  \emph{Prediche inutili}, Einaudi, Turin, 1956, pp.~1-12.
\item
  EINAUDI L., \emph{Corso di scienza della finanza tenuto dal
  Prof.~Luigi Einaudi nella R. Università di Torino e nella Università
  commerciale L. Bocconi di Milano}, La Riforma Sociale, Turin, 1916.
\item
  EINAUDI L., \emph{Il sistema tributario italiano}, 2nd ed., La Riforma
  Sociale, Turin, 1933.
\item
  EINAUDI L., \emph{La terra e l'imposta}, Einaudi, Turin, 1942.
\item
  EINAUDI L., \emph{Miti e paradossi della giustizia tributaria},
  Einaudi, Turin, 1940.
\item
  EINAUDI L., \emph{Saggi sul risparmio e l'imposta}, Einaudi, Turin,
  1941.
\item
  EPSTEIN J. M., \emph{Agent-Based Computational Models and Generative
  Social Science}, in \emph{Complexity}, 1999, vol.~4, no. 5, pp.~41-60,
  DOI:
  10.1002/(SICI)1099-0526(199905/06)4:5\textless41::AID-CPLX9\textgreater3.0.CO;2-F.
\item
  FATTORE G., \emph{La valutazione delle politiche: una breve guida
  metodologica}, in ADINOLFI P., URICCHIO A. F. (eds.), \emph{Valutare
  la complessità. La performance nelle organizzazioni pubbliche},
  Carocci, Rome, 2025, pp.~179 ff.
\item
  FROSINI V., \emph{Cibernetica, diritto e società}, Edizioni di
  Comunità, Milan, 1968.
\item
  FROSINI V., \emph{Telematica e informatica giuridica}, in
  \emph{Enciclopedia del diritto}, vol.~XLIV, Giuffrè, Milan, 1992,
  pp.~60-82.
\item
  GALETTA D.-U., CAPPIELLO B., \emph{Il Regolamento UE sull'intelligenza
  artificiale e le amministrazioni pubbliche: basi giuridiche, obiettivi
  e limiti di un esperimento normativo dagli esiti incerti}, in URICCHIO
  A. F., CALDAROLA C. (eds.), \emph{L'intelligenza artificiale tra
  regolazione ed esperienze applicative. AI Act (regolamento UE) e
  disegno di legge governativo}, Cacucci, Bari, 2025, pp.~59-86.
\item
  GALLO F., \emph{Cyberspazio, diritti e tutele}, in URICCHIO A. F.,
  CALDAROLA C. (eds.), \emph{L'intelligenza artificiale tra regolazione
  ed esperienze applicative. AI Act (regolamento UE) e disegno di legge
  governativo}, Cacucci, Bari, 2025, pp.~15-29.
\item
  GRECO M., \emph{Il diritto costituzionale dell'ambiente dopo la
  riforma: alcune conferme e qualche (inattesa) novità nella sentenza
  della Corte costituzionale n.~105/2024}, in \emph{Diritti comparati},
  25 July 2024.
\item
  GROVES D. G., LEMPERT R. J., \emph{A New Analytic Method for Finding
  Policy-Relevant Scenarios}, in \emph{Global Environmental Change},
  2007, vol.~17, no. 1, pp.~73-85, DOI: 10.1016/j.gloenvcha.2006.11.006.
\item
  GUSTON D. H., \emph{Understanding ``Anticipatory Governance''}, in
  \emph{Social Studies of Science}, 2014, vol.~44, no. 2, pp.~218-242,
  DOI: 10.1177/0306312713508669.
\item
  HAASNOOT M., KWAKKEL J. H., WALKER W. E., TER MAAT J., \emph{Dynamic
  Adaptive Policy Pathways: A Method for Crafting Robust Decisions for a
  Deeply Uncertain World}, in \emph{Global Environmental Change}, 2013,
  vol.~23, no. 2, pp.~485-498, DOI: 10.1016/j.gloenvcha.2012.12.006.
\item
  HANCHER L., MORAN M., \emph{Organizing Regulatory Space}, in BALDWIN
  R., SCOTT C., HOOD C. (eds.), \emph{A Reader on Regulation}, Oxford
  University Press, Oxford, 1998, pp.~148 ff.; originally in HANCHER L.,
  MORAN M. (eds.), \emph{Capitalism, Culture and Economic Regulation},
  Clarendon Press, Oxford, 1989.
\item
  HECKEL R., \emph{Graph Transformation in a Nutshell}, in
  \emph{Electronic Notes in Theoretical Computer Science}, 2006,
  vol.~148, no. 1, pp.~187-198, DOI: 10.1016/j.entcs.2005.12.018.
\item
  HOBBES T., \emph{Leviathan}, Latin version of 1668, in \emph{Opera
  philosophica quae Latine scripsit omnia}.
\item
  INGLESE M., \emph{Il regolamento sull'intelligenza artificiale come
  atto per il completamento e il buon funzionamento del mercato
  interno?}, in \emph{Quaderni AISDUE}, special issue no. 2/2024, pp.~1
  ff.
\item
  ISRAELI N., GOLDENFELD N., \emph{Computational Irreducibility and the
  Predictability of Complex Physical Systems}, in \emph{Physical Review
  Letters}, 2004, vol.~92, art. 074105, DOI:
  10.1103/PhysRevLett.92.074105.
\item
  LEMPERT R. J., GROVES D. G., POPPER S. W., BANKES S. C., \emph{A
  General, Analytic Method for Generating Robust Strategies and
  Narrative Scenarios}, in \emph{Management Science}, 2006, vol.~52, no.
  4, pp.~514-528, DOI: 10.1287/mnsc.1050.0472.
\item
  LEMPERT R. J., POPPER S. W., BANKES S. C., \emph{Shaping the Next One
  Hundred Years: New Methods for Quantitative, Long-Term Policy
  Analysis}, RAND Corporation, Santa Monica, 2003.
\item
  LESSIG L., \emph{Code Is Law: On Liberty in Cyberspace}, in
  \emph{Harvard Magazine}, vol.~102, no. 3, January-February 2000,
  pp.~37 ff.
\item
  PALUMBO M., \emph{Il processo di valutazione. Decidere, programmare,
  valutare}, FrancoAngeli, Milan, 2001.
\item
  PASCUZZI G., \emph{Il diritto dell'era digitale. Tecnologie
  informatiche e regole privatistiche}, Il Mulino, Bologna, 2002.
\item
  PASTORE B., \emph{Tecnologie emergenti, incertezze della scienza,
  regolamentazione giuridica}, in \emph{Teoria e Critica della
  Regolazione Sociale}, 2020, pp.~97 ff.
\item
  PENNASILICO M., review of CATERINI E., \emph{Sostenibilità e
  ordinamento civile. Per una riproposizione della questione sociale},
  ESI, Naples, 2018, in \emph{Rassegna di diritto civile}, 2018,
  pp.~1511 ff.
\item
  PENNASILICO M., \emph{La «sostenibilità ambientale» nella dimensione
  civil-costituzionale: verso un diritto dello «sviluppo umano ed
  ecologico»}, in \emph{Rivista quadrimestrale di diritto
  dell'ambiente}, no. 3/2020, pp.~4 ff.
\item
  PEPE V., \emph{Fare ambiente. Teorie e modelli giuridici di sviluppo
  sostenibile}, FrancoAngeli, Milan, 2008.
\item
  PEPE V., \emph{Le «nuove frontiere» del costituzionalismo ambientale:
  prospettive emergenti nel diritto comparato}, in \emph{DPCE Online},
  vol.~58, special issue 2/2023, pp.~963 ff., DOI:
  10.57660/dpceonline.2023.1929.
\item
  POLI S., \emph{Il rafforzamento della sovranità tecnologica europea e
  il problema delle basi giuridiche}, in \emph{I Post di AISDUE}, III
  (2021), section «Atti convegni AISDUE», no. 5, 20 December 2021,
  pp.~69 ff.
\item
  POPPER S. W., \emph{Robust Decision Making and Scenario Discovery in
  the Absence of Formal Models}, in \emph{Futures \& Foresight Science},
  2019, vol.~1, e22, DOI: 10.1002/ffo2.22.
\item
  PORENA D., \emph{Il principio della sostenibilità. Contributo allo
  studio di un programma costituzionale di solidarietà
  intergenerazionale}, Giappichelli, Turin, 2017.
\item
  RANCHORDÁS S., \emph{Experimental Regulations and Regulatory
  Sandboxes: Law Without Order?}, in \emph{Law and Method}, 2021, DOI:
  10.5553/REM/.000064.
\item
  REIDENBERG J. R., \emph{Lex Informatica: The Formulation of
  Information Policy Rules Through Technology}, in \emph{Texas Law
  Review}, 1998, vol.~76, no. 3, pp.~553-593.
\item
  RIVOSECCHI G., \emph{La garanzia costituzionale della copertura
  finanziaria come vincolo intertemporale alla spesa pubblica nella
  (limitata) prospettiva della tutela delle generazioni future}, in
  BIFULCO R., D'ALOIA A. (eds.), \emph{Un diritto per il futuro. Teorie
  e modelli dello sviluppo sostenibile e della responsabilità
  intergenerazionale}, Jovene, Naples, 2008, pp.~475 ff.
\item
  RUHL J. B., \emph{Complexity Theory as a Paradigm for the Dynamical
  Law-and-Society System: A Wake-Up Call for Legal Reductionism and the
  Modern Administrative State}, in \emph{Duke Law Journal}, 1996,
  vol.~45, pp.~849 ff.
\item
  RUHL J. B., \emph{Law's Complexity: A Primer}, in \emph{Georgia State
  University Law Review}, 2008, vol.~24, pp.~885 ff.
\item
  SABEL C. F., ZEITLIN J., \emph{Experimentalism in the EU: Common
  Ground and Persistent Differences}, in \emph{Regulation \&
  Governance}, 2012, vol.~6, no. 3, pp.~410-426, DOI:
  10.1111/j.1748-5991.2012.01157.x.
\item
  SABEL C. F., ZEITLIN J., \emph{Experimentalist Governance}, in
  LEVI-FAUR D. (ed.), \emph{The Oxford Handbook of Governance}, Oxford
  University Press, Oxford, 2012, DOI:
  10.1093/oxfordhb/9780199560530.013.0012.
\item
  SANDULLI M. A., APERIO BELLA F., LORENZONI L., POSTERARO N., SINISI
  M., \emph{Shaping the Future of Health Law: Challenges for Public
  Law}, paper published in \emph{federalismi.it}, 17 November 2021.
\item
  SCHREPEL T., \emph{Adaptive Regulation}, in \emph{European Journal of
  Risk Regulation}, 2026, pp.~374 ff., DOI: 10.1017/err.2026.10087.
\item
  TAFARO L., \emph{Diritti umani oggi: sviluppo sostenibile e
  generazioni future}, in CANÇADO TRINDADE A. A., BARROS LEAL C. (eds.),
  \emph{Diritti umani e ambiente. VI Corso brasiliano interdisciplinare
  in diritti umani}, Fortaleza, 2017, pp.~43-73.
\item
  TORRETTA P., \emph{Responsabilità intergenerazionale e procedimento
  legislativo. Soggetti, strumenti e procedure di positivizzazione degli
  interessi delle generazioni future}, in BIFULCO R., D'ALOIA A. (eds.),
  \emph{Un diritto per il futuro. Teorie e modelli dello sviluppo
  sostenibile e della responsabilità intergenerazionale}, Jovene,
  Naples, 2008, pp.~699 ff.
\item
  VAN DEVENTER O. et al., \emph{Towards European Standards for Quantum
  Technologies}, in \emph{EPJ Quantum Technology}, 2022, vol.~9, art.
  33, DOI: 10.1140/epjqt/s40507-022-00150-1.
\item
  VAN OOIJEN C., UBALDI B., WELBY B., \emph{A data-driven public sector:
  Enabling the strategic use of data for productive, inclusive and
  trustworthy governance}, OECD Working Papers on Public Governance, no.
  33, OECD Publishing, Paris, 2019, DOI: 10.1787/09ab162c-en.
\item
  WOLFRAM S., \emph{A New Kind of Science}, Wolfram Media, Champaign,
  2002.
\item
  WOLFRAM S., \emph{What Is Ruliology?}, 12 January 2026, available on
  the author's website.
\end{itemize}

\subsection{Institutional Documents and Official
Sources}\label{institutional-documents-and-official-sources}

\begin{itemize}
\item
  EUROPEAN COMMISSION, \emph{Commission Invites Contributions to Shape
  Future EU Quantum Act}, press release of 31 October 2025, updated on
  24 November 2025.
\item
  EUROPEAN COMMISSION, \emph{EU Quantum Act --- Call for Evidence for an
  Impact Assessment}, document Ares(2025)9260984, 2025.
\item
  EUROPEAN COMMISSION, \emph{Quantum Europe Strategy: Quantum Europe in
  a Changing World}, COM(2025)363 final, Brussels, 2 July 2025.
\item
  COUNCIL OF THE EUROPEAN UNION, \emph{Russian Hybrid Threats: EU
  Sanctions}, official source on Russian hybrid and cyber activities,
  version consulted on 30 August 2026.
\item
  JOINT RESEARCH CENTRE, LEWIS A. M. et al., \emph{Future Directions for
  Quantum Technology in Europe: An Analysis of Policy Questions},
  JRC141050, Publications Office of the European Union, Luxembourg,
  2025, DOI: 10.2760/0891087.
\item
  OECD, \emph{2023 OECD Digital Government Index: Results and Key
  Findings}, OECD Public Governance Policy Papers, OECD Publishing,
  Paris, 2024, DOI: 10.1787/1a89ed5e-en.
\item
  OECD, \emph{A Quantum Technologies Policy Primer}, OECD Science,
  Technology and Industry Policy Papers, no. 371, OECD Publishing,
  Paris, 2025, DOI: 10.1787/fd1153c3-en.
\item
  OECD, \emph{Framework for Anticipatory Governance of Emerging
  Technologies}, OECD Science, Technology and Industry Policy Papers,
  no. 165, OECD Publishing, Paris, 2024, DOI: 10.1787/0248ead5-en.
\item
  OECD, \emph{Governing with Artificial Intelligence: The State of Play
  and Way Forward in Core Government Functions}, OECD Publishing, Paris,
  2025, DOI: 10.1787/795de142-en.
\item
  OECD, \emph{Regulatory Experimentation: Moving Ahead on the Agile
  Regulatory Governance Agenda}, OECD Public Governance Policy Papers,
  no. 47, OECD Publishing, Paris, 2024, DOI: 10.1787/f193910c-en.
\item
  OECD, \emph{Tax Administration 2025: Comparative Information on OECD
  and Other Advanced and Emerging Economies}, OECD Publishing, Paris,
  2025, DOI: 10.1787/cc015ce8-en.
\item
  OECD, \emph{The Path to Becoming a Data-Driven Public Sector}, OECD
  Digital Government Studies, OECD Publishing, Paris, 2019, DOI:
  10.1787/059814a7-en.
\item
  PRESIDENCY OF THE COUNCIL OF MINISTERS --- INTELLIGENCE SYSTEM FOR THE
  SECURITY OF THE REPUBLIC, \emph{Relazione sulla politica
  dell'informazione per la sicurezza 2024}, Rome, 2025.
\item
  HOLY SEE, LEO XIV, \emph{Magnifica Humanitas}, 15 May 2026.
\item
  THE WHITE HOUSE, \emph{Winning the Race: America's AI Action Plan},
  Washington, 23 July 2025.
\end{itemize}

\end{document}